\documentclass{article}

\PassOptionsToPackage{numbers, compress}{natbib}

\usepackage[pdftex]{graphicx}
\usepackage[final]{neurips_2024}

\usepackage[utf8]{inputenc} % allow utf-8 input
\usepackage[T1]{fontenc}    % use 8-bit T1 fonts
\usepackage{hyperref}       % hyperlinks
\usepackage{url}            % simple URL typesetting
\usepackage{booktabs}       % professional-quality tables
\usepackage{diagbox}
\usepackage{amsfonts}       % blackboard math symbols
\usepackage{nicefrac}       % compact symbols for 1/2, etc.
\usepackage{enumitem}
\usepackage{microtype}      % microtypography
\usepackage[table,xcdraw]{xcolor}
\usepackage{colortbl}
\usepackage{multirow}
\usepackage{comment}
\usepackage{amsmath}
\usepackage{wrapfig}
\usepackage{makecell}
\usepackage{arydshln}

\usepackage{subcaption}  

\definecolor{MyDarkBlue}{rgb}{0,0.08,1}
\definecolor{MyDarkGreen}{rgb}{0.02,0.6,0.02}
\definecolor{MyDarkRed}{rgb}{0.8,0.02,0.02}
\definecolor{MyDarkOrange}{rgb}{0.40,0.2,0.02}
\definecolor{MyPurple}{RGB}{111,0,255}
\definecolor{MyRed}{rgb}{1.0,0.0,0.0}
\definecolor{MyGold}{rgb}{0.75,0.6,0.12}
\definecolor{MyDarkgray}{rgb}{0.66, 0.66, 0.66}

\newcommand{\methodname}{LeapAD}
\newcommand{\fastsystem}{Heuristic Process}
\newcommand{\slowsystem}{Analytic Process}

\hypersetup{colorlinks=true,linkcolor=red,citecolor=green,urlcolor=magenta}
\usepackage[font={small}]{caption}
\title{Continuously Learning, Adapting, and Improving:\\ A Dual-Process Approach to Autonomous Driving}

\author{%
  Jianbiao Mei$^{1,2,*}$
  \quad
  Yukai Ma$^{1,2,*}$ 
  \quad
  Xuemeng Yang$^{2}$
  \quad
  Licheng Wen$^{2}$
  \quad 
  Xinyu Cai$^{2}$ \\
  \textbf{Xin Li$^{2,4}$}
  \quad
  \textbf{Daocheng Fu$^{2}$}
  \quad
  \textbf{Bo Zhang$^{2}$}
  \quad
  \textbf{Pinlong Cai$^{2}$}
  \quad
  \textbf{Min Dou$^{2}$} \\
  \textbf{Botian Shi$^{2,\dag}$}
  \quad
  \textbf{Liang He$^{3}$}
  \quad
  \textbf{Yong Liu$^{1,\dag}$}
  \quad
  \textbf{Yu Qiao$^{2}$} \\
  $^1$ Zhejiang University
  $^2$ Shanghai Artificial Intelligence Laboratory \\
  $^3$ East China Normal University
  $^4$ Shanghai Jiao Tong University
}
\begin{document}
\maketitle

\renewcommand{\thefootnote}{\relax}
\footnotetext{* equal contribution, $\dag$ corresponding author}
\renewcommand{\thefootnote}{\arabic{footnote}}

\begin{abstract}
Autonomous driving has advanced significantly due to sensors, machine learning, and artificial intelligence improvements. However, prevailing methods struggle with intricate scenarios and causal relationships, hindering adaptability and interpretability in varied environments. 
To address the above problems, we introduce \textbf{\methodname}, a novel paradigm for autonomous driving inspired by the human cognitive process. Specifically, {\methodname} emulates human attention by selecting critical objects relevant to driving decisions, simplifying environmental interpretation, and mitigating decision-making complexities. Additionally, {\methodname} incorporates an innovative dual-process decision-making module, which consists of an {\slowsystem} (System-II) for thorough analysis and reasoning, along with a {\fastsystem} (System-I) for swift and empirical processing. 
The {\slowsystem} leverages its logical reasoning to accumulate linguistic driving experience, which is then transferred to the {\fastsystem} by supervised fine-tuning.
Through reflection mechanisms and a growing memory bank, {\methodname} continuously improves itself from past mistakes in a closed-loop environment.
Closed-loop testing in CARLA shows that {\methodname} outperforms all methods relying solely on camera input, requiring 1-2 orders of magnitude less labeled data. Experiments also demonstrate that as the memory bank expands, the {\fastsystem} with only 1.8B parameters can inherit the knowledge from a GPT-4 powered {\slowsystem} and achieve continuous performance improvement.
Project page: \url{https://pjlab-adg.github.io/LeapAD/}.
\end{abstract}

\section{Introduction}
% *********************************************************************************
Since the early 21st century, starting with the DARPA Grand Challenge~\cite{thrun2006stanley}, humanity has explored replacing human drivers with computer algorithms. Over the past two decades, advancements in sensor technology, machine learning, and artificial intelligence have propelled the evolution of self-driving technology. 
Recent data-driven approaches achieved considerable success, as evidenced by new vehicle models featuring intelligent driving assistance and the commercial operation of L4 robotaxis in several cities~\cite{yin2021center,li2022bevformer,liu2023bevfusion}.
However, these methods depend heavily on diverse training data distributions, resulting in a superficial understanding of underlying semantics and potential misconceptions in complex situations.
This is because data-driven approaches primarily perform induction on observed patterns without the capability for deduction, thus constraining their performance to the coverage of the annotated data.
Therefore, there is an urgent need for a system capable of reasoning about unseen scenarios and utilizing knowledge in a human cognition manner.

The latest advancements in Large Language Models (LLMs) and Vision Language Models (VLMs), noted for their embedded world knowledge and robust explanatory and reasoning capabilities, have captured the interest of researchers~\cite{wayve2023lingo,ma2023dolphins,li2023towards,zhou2024embodied}. For example, in the autonomous driving field, some knowledge-based methods ~\cite{shao2023lmdrive, mao2023language, yuan2024rag, tian2024drivevlm} employ LLMs and VLMs as the driving agents.
However, these methods perform open-loop testing, which merely evaluates errors between model output and the ground truth from datasets, failing to reflect the dynamic interactions between ego car and the real-world environment~\cite{li2023ego}. Consequently, they are often inadequate to effectively assess the responsiveness and adaptability of driving agents.

In fact, human learning to drive involves a continuous interaction and exploration process within closed-loop environments, where drivers make decisions based on the surroundings and receive feedback accordingly.
As per the dual-process theory~\cite{kahneman2011fast, evans2013dual, wason1974dual}, human intelligence operates on two levels: 1) \emph{\fastsystem} (\emph{System-I}), which is automatic, quick, empirical, and domain-specific; and 2) \emph{\slowsystem} (\emph{System-II}), which is rational, slow, and excels in logical reasoning and creativity across various domains. This dual-process thinking is evident in the progression from novice to experienced driver. Initially, individuals rely heavily on common sense due to their lack of driving experience. Through training, they develop driving skills via a closed-loop learning process involving continuous trial and error, along with rational analysis (\slowsystem) to evaluate their behavior. These skills become internalized over time, forming muscle memory that enables quick, instinctive reactions in familiar driving scenarios (\fastsystem). Even after obtaining driver's license, individuals continue to gain experience and learn from accidents to enhance driving skills.

To this end, we develop a dual-process closed-loop autonomous driving system that is continuously \textbf{le}arning, \textbf{a}dapting and im\textbf{p}roving, named \textbf{\methodname}.
Similar to the human attention mechanism, the scene understanding module in {\methodname} mainly focuses on critical objects that may affect driving decisions, simplifying the environmental description and the decision-making process.
Following such scene understanding, we develop a dual-process decision-making module that emulates human cognitive processes, featuring a {\fastsystem} and an {\slowsystem} \cite{epstein2003cognitive,evans2013dual}. 
Through a closed-loop setup, {\slowsystem} accumulates experience and builds a transferable memory bank of high-quality driving decisions.
The knowledge can be adapted to various scenarios and then transferred to the lightweight model in {\fastsystem} through supervised fine-tuning (SFT).
The {\fastsystem} is employed for closed-loop decision-making using a few-shot strategy. When traffic accidents occur, the {\slowsystem} intervenes to analyze these incidents and update the memory bank, enabling the system to continuously improve through self-reflection.
The main contributions of our work are summarized as follows:
\vspace{-2mm}
\begin{itemize}[align=right,itemindent=0em,labelsep=2pt,labelwidth=1em,leftmargin=*,itemsep=0em] 
\item We develop an innovative closed-loop autonomous driving approach that emulates the critical object attention mechanisms and the learning processes observed in human driving behavior.
\item We propose a dual-process decision-making module inspired by human cognition theory. In the absence of human involvement, our approach enables the fast, empirical {\fastsystem} to inherit the capabilities of the slow, rational {\slowsystem} in a self-supervised manner.
\item {\methodname} utilizes the {\slowsystem} and a reflection mechanism to accumulate a transferable memory bank, enabling the system to achieve continuous learning and generalization capabilities in a closed-loop driving environment.
\item  Extensive experiments in CARLA show that {\methodname} not only outperforms all other methods relying solely on camera input, but also achieves this with 1-2 orders of magnitude less annotated data.
\end{itemize}
\vspace{-1em}
\section{Related Works}

\vspace{-0.15cm}
\subsection{Large Vision Language Models}
Inspired by the successful deployment of Large Language Models (LLMs) like LLaMAs~\cite{touvron2023llama, touvron2023llama2} and Vicuna~\cite{chiang2023vicuna}, a plethora of Vision Language Models (VLMs) ~\cite{alayrac2022flamingo, chen2022pali, li2023mimic, li2023videochat, li2023monkey, zhang2023video, wang2023cogvlm, bai2023qwen, chen2023internvl} has emerged to broaden their applicability to multi-modal understanding. Various models, such as BLIP2~\cite{li2023blip} which utilizes the Q-former, Flamingo~\cite{alayrac2022flamingo} leveraging a perceiver resampler, and LLaVA~\cite{liu2024visual} alongside MiniGPT-4~\cite{zhu2023minigpt} that incorporate instruction tuning, have been innovated to enhance feature alignment, few-shot learning, and create versatile visual agents. Moreover, models like Qwen-VL~\cite{bai2023qwen} with its three-stage training, and InternVL~\cite{chen2023internvl}'s image-text alignment method, assist in achieving advanced multi-lingual and fine-grained visual comprehension.
The rise of VLM and visual-language-action (VLA) models has injected the vitality of autonomous driving, presenting researchers with new opportunities. 

\vspace{-0.15cm}
\subsection{Empowering Autonomous Driving with Foundation Models}
Recent work \cite{fu2024drive, wen2023dilu, shao2023lmdrive, mao2023language, yuan2024rag, wen2024on} explores the use of large foundation models in autonomous driving, leveraging their embedded world knowledge and powerful interpretation and reasoning capabilities.
For understanding driving scenarios, a series of datasets and benchmarks~\cite{sima2023drivelm, marcu2023lingoqa, ma2023dolphins, nie2023reason2drive} have been proposed.
To improve the interpretability of autonomous driving, LMDrive~\cite{shao2023lmdrive} and DriveMLM~\cite{wang2023drivemlm} use LLMs to generate human-instructed decisions in the simulated environment, which is data-dependent and hard to adapt to the real world.
Agent-Driver~\cite{mao2023language} adopts LLM agent for planning, which is less efficient due to excess environmental data.
ELM~\cite{zhou2024embodied} introduces a vision-language model tailored for embodied understanding within driving scenarios.
RAG-Driver~\cite{yuan2024rag} improves driving interpretation and signal prediction by integrating retrieval augmentation and in-context learning. 
Recent DriveVLM-Dual~\cite{tian2024drivevlm} integrates VLM and data-driven planning pipelines, providing solutions for deployment.
Contrary to the mentioned techniques, our {\methodname} draws from attention mechanisms and observational learning and decision-making in human driving. It utilizes a memory bank for experience storage and replay in a closed-loop scenario, enabling continuous learning via memory and reflection mechanisms.

\vspace{-0.2cm}
\subsection{From Data-Driven to Knowledge-driven Autonomous Driving}
While the prevailing data-driven approaches~\cite{yin2021center,li2022bevformer,liu2023bevfusion, jia2023think, shao2023safety, chitta2022transfuser, wu2022trajectory, hu2023planning,chitta2021neat, zhang2021end, codevilla2019exploring, chen2020learning, jiang2023vad} have led to success in both academia and industry in past decades, allowing autonomous driving technology to be used in people's daily lives.
However, these methods are limited to the distribution of training data and frequently encounter adaptability issues and long-tail challenges when expanding across different  areas~\cite{peng2023tong,gildert2023building}. 
On the other hand, human drivers possess a deep common sense of understanding the world, which enables them to adapt to unexpected scenarios.
This highlights the need for a shift to the knowledge-driven approaches, which involve using empirical reasoning and induction to learn from the environment~\cite{liu2020overview,dou2023towards}, and updating insights to develop specialized skills~\cite{zhang2023toward,xi2023rise}.
Knowledge-driven methods acquire general knowledge rather than merely implementing predefined human rules or abstracting characteristics from collected data in specific domains~\cite{li2023towards}. These approaches enhance performance, interpretability, and safety by integrating human-like logic into AI systems, particularly in managing complex traffic scenarios.
In the era of foundation models, the advanced reasoning and knowledge application capabilities exhibited by LLMs and VLMs have proven highly effective for complex tasks such as understanding, reasoning, and decision-making within the domain of autonomous driving~\cite{fu2024drive,liu2023can,cui2024survey}. These foundation models have embedded world knowledge and robust explanatory and reasoning capabilities through extensive training on diverse datasets and captured the researchers' interest~\cite{wayve2023lingo,ma2023dolphins,li2023towards,zhou2024embodied}.

\vspace{-0.15cm}

% *********************************************************************************

% *********************************************************************************
\section{Methodology}

\subsection{Overview}
\begin{figure}[tbp]
    \centering
    \includegraphics[width=0.95\linewidth]{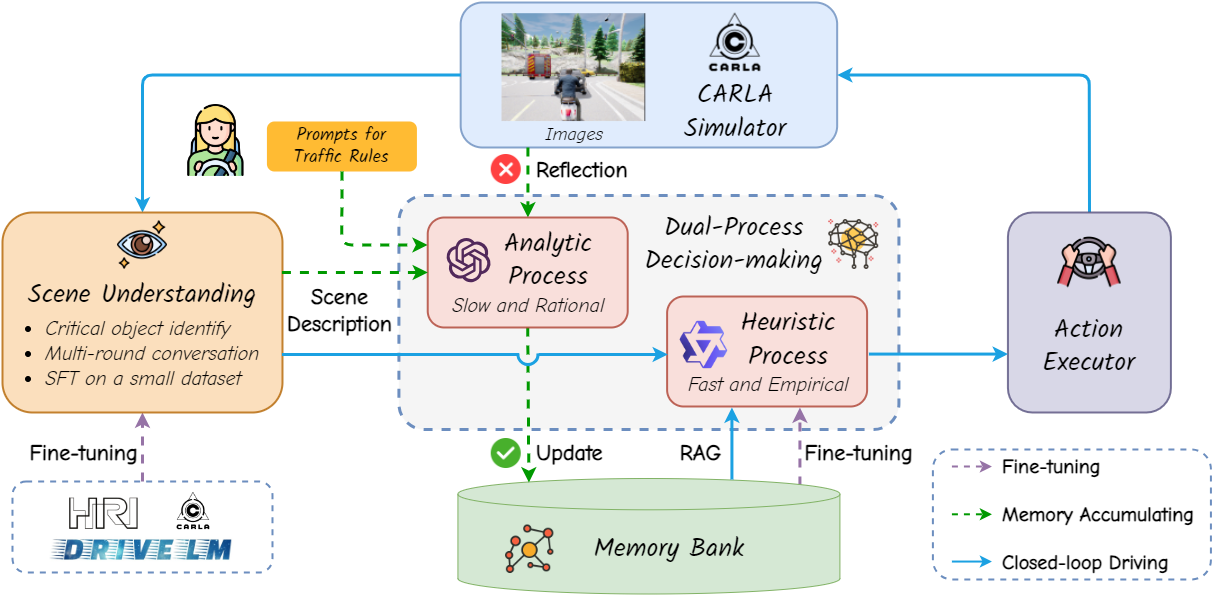}
    \caption{The detailed architecture of our proposed {\methodname}. The scene understanding module analyzes surrounding images and provides descriptions of critical objects that may influence driving decisions. These scenario descriptions are then fed into the dual-process decision module, which drives reasoning and decision-making. The generated decisions are then transmitted to action executors, where they are converted into control signals for interaction with the simulator. 
    The {\slowsystem} then uses an LLM to accumulate experience in driving analysis and decision-making, conducting reflections on accidents. The experience is stored in the memory bank and transferred to a lightweight language model, forming our {\fastsystem} for quick responses and continuous learning.
    }
    \label{fig:method}
    \vspace{-15pt}
\end{figure}

In this section, we introduce how we design our anthropomorphic closed-loop autonomous driving system, {\methodname}. Figure~\ref{fig:method} illustrates that {\methodname} consists of three main components: the VLM for scene understanding (Section~\ref{method:vlm}), the dual-process decision-making module comprising the {\slowsystem} (Section~\ref{method:slow}) and the {\fastsystem} (Section~\ref{method:fast}), along with the action executor for low-level control (Appendix~\ref{sec:low-level control}). 
In the CARLA simulator, {\methodname} utilizes VLM to process the surrounding images and generate descriptions of critical objects.
These scene descriptions are then fed into the dual-process decision-making module in order to derive scene reasoning and driving decisions. Finally, these high-level decisions are forwarded to the action executor, translated into control signals, and interact with the simulator.

In closed-loop driving environments, the fine-tuned lightweight model in {\fastsystem} is used to perform quick, empirical decisions with the transferable experience in the memory bank.
And when the {\fastsystem} encounters accidents, the {\slowsystem} intervenes.  The {\slowsystem} exploits LLMs to analyze traffic accidents, leveraging its embedded world knowledge, particularly its understanding of traffic rules. It then generates corrected, high-quality driving experiences, enriching the memory bank and enabling continuous learning for the entire system.

\subsection{Scene Understanding with VLM} \label{method:vlm}
Human drivers typically focus on critical elements surrounding the vehicle to prevent information overload, enhance reaction time, and minimize cognitive load. This approach helps improve driving concentration and reduces accident probabilities.
Inspired by such a mechanism, the scene understanding module in {\methodname} is designed to selectively identify critical objects, simplifying the description of the surrounding environment and reducing the load on decision-making processes.

Specifically, since off-the-shelf foundation VLMs lack domain-specific knowledge in the driving domain, we perform SFT and prompt the VLMs to output the linguistic descriptions of the objects that may influence subsequent driving decisions. 
The description of these critical objects includes their semantic, spatial, motion attributes, and behavioral reasoning. Integrating these aspects promotes a comprehensive understanding of the environment, which can ensure safety and adaptability in complex and dynamic driving environments.
For a specific driving scene, the descriptions generated by VLM can be expressed as $D=\{A_{s, i}, A_{l, i}, A_{m, i}, C_{r, i}\}_{i=0}^{N-1}$, where $N$ denotes the number of the critical objects.
For each critical object $O_i$, the description contains: 
\emph{\romannumeral 1 )} the semantic attribute $A_s$ describes its semantic category, usually important traffic participants (e.g., vehicles and cyclists) and infrastructure (e.g., traffic lights and stop signs). \emph{\romannumeral 2 )} The spatial attribute $A_l$ indicates its bounding box, the lane it locates, and the distance from the ego car, which are important for safety and collision avoidance. 
\emph{\romannumeral 3 )} The motion attribute $A_m$ refers to the motion direction of the object. \emph{\romannumeral 4 )} Behavioral reasoning $C_r$ describes why the object is critical and how it influences the driving decision of the ego car. For example, when the ego car goes straight, the stop sign on the right side is of high importance because it indicates the need to stop at the intersection.
We provide an example to further illustrate the descriptions of critical objects in the driving scene, as shown in Figure \ref{fig:data_illustrate} in Appendix~\ref{sec:data_anno}. 
Notably, the VLM not only excels in simulated environments but also demonstrates robust performance in real-world scenarios.

\begin{figure}
    \centering
    \includegraphics[width=\linewidth]{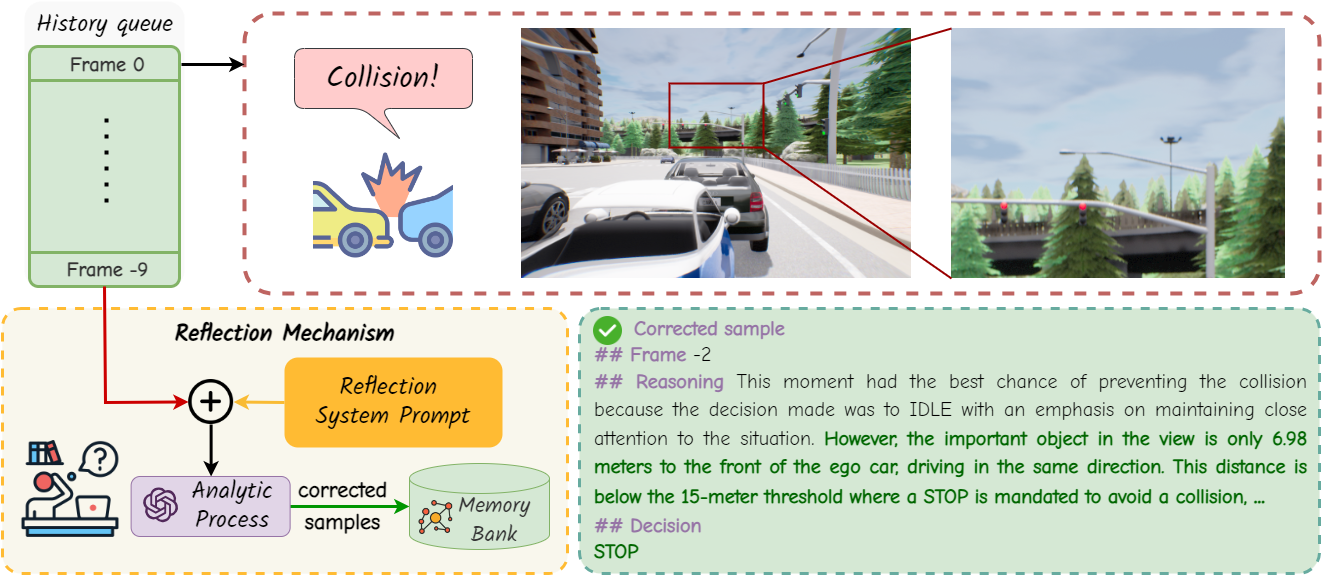}
    \caption{Detailed procedure of the reflection mechanism. When {\fastsystem} encounters traffic accidents, the {\slowsystem} intervenes, analyzing historical frames to pinpoint errors and provide corrected samples. These corrected samples are then integrated into the memory bank to facilitate continuous learning.}
    \label{fig:reflection}
    \vspace{-15pt}
\end{figure}

\subsection{\slowsystem} \label{method:slow}
Based on the scene descriptions provided by the VLM, we design the {\slowsystem} to imitate the rational thinking of a human driver. The {\slowsystem} relies on logical reasoning, employing rational thinking to analyze complex situations and make safe driving decisions.
The LLMs, through their extensive pre-training on diverse datasets, have encapsulated vast amounts of world knowledge, equipping them with the ability to handle intricate problems with nuanced understanding and reasoning~\cite{wen2023dilu}. 
This capability aligns with the requirements of the {\slowsystem} in driving scenarios, where decisions must be made based on deep analysis and contextual understanding of the environment. 
Our {\slowsystem} harnesses the power of LLMs, leveraging its world knowledge to understand the scene descriptions and perform high-quality driving analysis and decisions. We empirically found that prompting LLMs with specific traffic rules provided in Appendix~\ref{sec:prompt} further improves safety and is more reliable for on-road scenarios.

Furthermore, we integrate the VLM and the {\slowsystem} to run closed-loop experiments and collect the high-quality decision-making processes and results generated by the {\slowsystem} as "experience" in a memory bank.
The accumulated experience can be seamlessly transferred to the {\fastsystem}, facilitating it to react quickly based on experience when handling similar situations, as described in Section \ref{method:fast}.

\textbf{Reflection mechanism.} We also employ the {\slowsystem} to reflect on traffic accidents, as shown in Figure \ref{fig:reflection}.
Specifically, when VLM and {\fastsystem} run in a closed-loop driving scenario, any accident will trigger the reflection mechanism. During this procedure, the scene description $D$, reasoning $R$, and decision $S$ of the preceding frames before the accident are forwarded to {\slowsystem}. It is then required to meticulously analyze the cause of the event, locate the error, and provide corrected reasoning and decisions.
The insights gained from the reflection procedure are further integrated into the memory bank, allowing the {\methodname} to continuously learn from failures and progressively lead to more informed and accurate decision-making in future driving scenarios.
Importantly, the experience in the memory bank has good transferability and generalization. It can be directly utilized by other lightweight models and easily generalized to different scenarios, as demonstrated in section \ref{ab:generalization}.

\subsection{\fastsystem} \label{method:fast}
While the {\slowsystem} can offer more precise driving reasoning and decisions due to its detailed analysis and careful consideration, the inherent slow processing causes duplicated and redundant effort, limiting its application in practical driving scenarios. 
In contrast, human drivers form muscle memory through repeated practice and experience, requiring less effort over time. 
To reflect this quick and empirical thinking pattern and facilitate practical application, we craft a {\fastsystem} in {\methodname} incorporating a lightweight language model. 
Specifically, we perform supervised fine-tuning (SFT) using the samples stored in the accumulated memory bank mentioned in Section \ref{method:slow} to distill knowledge into the lightweight language model. By this means, the {\fastsystem} achieves behavior adaption to various scenarios and runs much faster than {\slowsystem} {(about 5 times faster in our experiments)}. We empirically found that the lightweight model without SFT is unable to produce appropriate driving decisions.

\textbf{Few-shot Prompting.} Moreover, we perform few-shot prompting~\cite{wen2023dilu} to enhance the {\fastsystem}'s generalization ability for unseen scenes and mitigate hallucinations for more robust decisions. 
\begin{wrapfigure}{r}{0.52\textwidth}
    \centering
    \includegraphics[width=\linewidth]{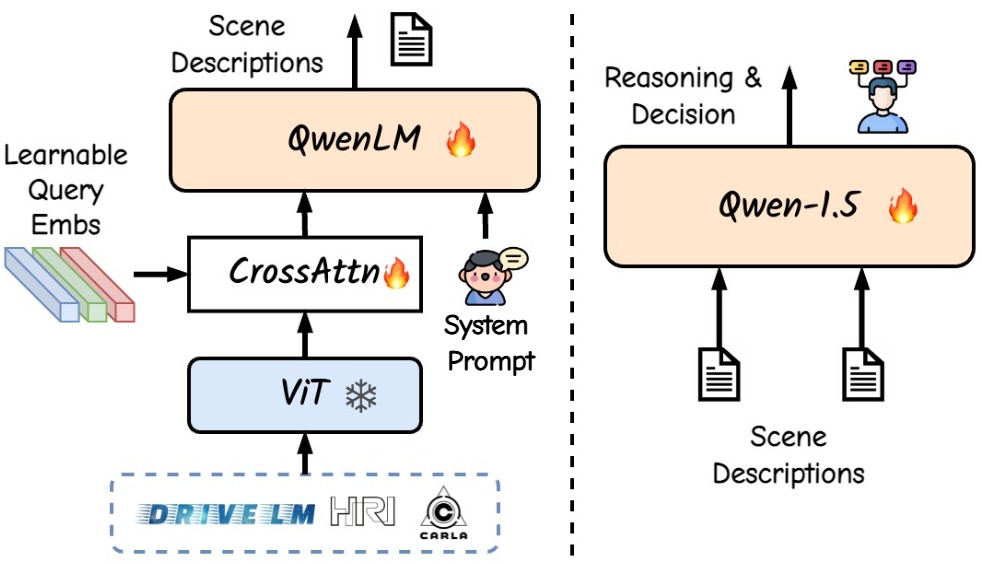}
    \caption{The illustration of the fine-tuning process. We fine-tune the VLM (Qwen-VL-7B) using 11K instruction-following data for scene understanding (left). Also, we utilize the collected samples in the memory bank to fine-tune Qwen-1.5 used in {\fastsystem}, as illustrated in the right part.}
    \label{fig:finetune}
    \vspace{0pt}
\end{wrapfigure}
Through such a mechanism, the {\fastsystem} can effectively leverage the experience and deep insights from the existing memory bank, improving the accuracy of future driving decisions. 
To facilitate the retrieval of similar driving scenes from the memory bank for few-shot prompting, we primarily rely on the embedding similarity between the current scene's descriptions and those stored in the memory bank. 
However, if directly calculating text similarity based on the original descriptions, the presence of redundant linguistic information in descriptions can complicate the differentiation between scenes.
Thus, we propose a novel scene encoding method to extract and encode compressed captions that comprise key elements such as the critical object's category, lane, and distance from the ego car. 
This approach streamlines the procedure of querying similar scenarios, enhancing retrieval efficiency and accuracy by prioritizing the most influential aspects for driving decisions of each scenario.
Afterward, the compressed captions are sent to a text encoder to encode the embedding vectors, which are expressed as follows:
\begin{equation}
    \textbf{e} = T_e(F_c(D)),
\end{equation} where $D$ is the scene description, $F_c$ denotes the compressing process, and $T_e$ indicates the text encoder.
Subsequently, the cosine similarity between the query embedding $\textbf{e}_q$ for the current scene and the embedding $\{\textbf{e}_i\}_{i=0}^{M-1}$ for the memory bank with the size of $M$ is computed by:
\begin{equation}
    s(\textbf{e}_q, \textbf{e}_i) = \frac{\textbf{e}_q\cdot\textbf{e}_i}{\|\textbf{e}_q\| \|\textbf{e}_i\|}.
\end{equation}
We select top-k samples with the highest similarity scores as queried scenes. The scene descriptions $\{D_i\}_{i=0}^{k-1}$, reasoning $\{R_i\}_{i=0}^{k-1}$, and decisions $\{S_i\}_{i=0}^{k-1}$ of k samples and the scene descriptions ${D}_c$ of the current scene are both fed into the {\fastsystem} for the final reasoning ${R}_c$ and decision ${S}_c$.

\section{Experiments}

\subsection{Data preparation}
\label{sec: exp-data}
\paragraph{Data for VLM.} 
\label{sec:exp-data-vlm}
We construct the instruct-following datasets for supervised fine-tuning of our VLM by integrating Rank2Tell~\cite{sachdeva2024rank2tell}, DriveLM~\cite{sima2023drivelm}, and data collected within CARLA~\cite{dosovitskiy2017carla}. To maintain consistency across all the datasets, we adopt a uniform standard reference format for critical objects as: \texttt{<ref>In \{camera view\}, \{properties\}</ref><box>\{coordinates\}</box>}. For each dataset, specific Q\&A pairs are created to suit their unique structures and contents.
The conversations are structured in a summary-elaboration manner, with the first question for the Rank2Tell and DriveLM datasets focusing on determining the number, semantic, and spatial attributes, such as the bounding box coordinates of key objects.
For the Rank2Tell dataset, we follow up by inquiring about the moving state, importance, and corresponding reasoning for each critical object. For the DriveLM dataset, we retain most of the original questions but eliminate redundant ones. We extract only $6\rm{K}$ frames of data from Rank2tell and DriveLM and organize them in the standardized format.
The data gathered from the CARLA simulator is exclusively dedicated to the closed-loop experiments detailed in Section~\ref{exp: close-loop}. A comprehensive training dataset of $5\rm{K}$ frames is collected from Town 01-04, 06, 07, and 10. To identify key objects within the scene, we design several automatic annotation rules, as detailed in Appendix \ref{sec:data_anno}. For clarity, we provide the data illustration in Figure \ref{fig:data_illustrate} in Appendix \ref{sec:data_anno}.

\paragraph{Data for {\fastsystem}.}
We leverage the integration of {\slowsystem} and VLM to accumulate experience within the closed-loop setup and save it in the memory bank for subsequent SFT and few-shot prompting of {\fastsystem}. Moreover, our approach incorporates dynamic updates to address errors encountered by the {\fastsystem}, as mentioned in the reflection mechanism outlined in Section \ref{method:slow}. The memory bank in our approach is configured to a default size of $9.0\rm{K}$, including samples collected from various towns (01-04, 06, 07, and 10).
It is worth noting that samples are obtained in a closed-loop environment without human involvement. Each sample consists of the scene descriptions $D$ depicted in Section \ref{method:vlm}, reasoning $R$, and decisions $S$. The reasoning $R$ explains the process of decision-making based on the scene descriptions $D$ and chain of thought. 

\subsection{{Implementation Details}}
\label{sec: exp-detail}
We employ Qwen-VL-7B~\cite{bai2023qwen} as the VLM for scene understanding, GPT-4 as the {\slowsystem} for rational and logic thinking, and Qwen1.5-1.8B~\cite{qwen} as {\fastsystem} for automatic and quick thinking in {\methodname}. We use the OpenAI embedding model as the text encoder $T_e$ to extract text embedding.
To fully excite our VLM's capabilities in autonomous driving, we perform SFT with the instruction-following data discussed in Section~\ref{sec:exp-data-vlm}. We utilize the AdamW optimizer \cite{loshchilov2017decoupled} with $\beta_1 = 0.9$ and $\beta_2 = 0.95$, coupled with a cosine decay of the learning rate, initially set to $1e^{-5}$. The batch size is set to 16, and the model is trained for 5 epochs on 8 A100 GPUs, requiring about 26 hours. The input image resolution is set at $448\times448$ pixels. 
For {\fastsystem}, we conduct SFT on Qwen1.5-1.8B for 5 epochs using samples stored in the memory bank, taking about 6 hours. The training hype-parameters are consistent with the training procedure of VLM. The detailed fine-tuning process is shown in Figure \ref{fig:finetune}. The dual-process decision module outputs meta-actions (e.g., "AC", "DC", "IDLE", "STOP") at a frequency of 2 HZ, which are further refined to control signals, as detailed in the Appendix~\ref{sec:low-level control}. Please refer to Appendix \ref{sec:reflection} for more details about the reflection mechanism and Appendix~\ref{sec:data_anno} for the performance of VLM on both simulated and real datasets.

\begin{table*}[t]
    \centering
    \small
    \renewcommand{\arraystretch}{1.1}
    \caption{Comparison of our {\methodname} with competitive methods on Town05 Short benchmark. 
    Notably, {\methodname} demonstrated superior performance with a smaller data footprint, outperforming other approaches. "DD" \& "KD" denote data-driven and knowledge-driven, respectively. "L" \& "C" indicate LiDAR and camera modalities.}
    \begin{tabular}{l|ccccc}
    \Xhline{1pt}
        {Method} & {Modality} & Type & Annotations & {DS $\uparrow$} & {RC $\uparrow$} \\ \Xhline{1pt}
        InterFuser~\cite{shao2023safety} & L+C & DD & $3\rm{M}$ & \textbf{94.95$\pm$1.91} & \textbf{95.19$\pm$2.57} \\ 
        TransFuser~\cite{chitta2022transfuser} & L+C & DD & $ 228\rm{K}$ & 54.52$\pm$4.29 & 78.41$\pm$3.75 \\
        \hline
        VAD~\cite{jiang2023vad} & C & DD & 228\rm{K} & 64.30 & 87.30 \\
        NEAT~\cite{chitta2021neat} & C & DD & $130\rm{K}$ & 58.70$\pm$4.11 & 77.32$\pm$4.91 \\
        Roach~\cite{zhang2021end} & C & DD & - & 65.26$\pm$3.63 & 88.24$\pm$5.16 \\
        WOR~\cite{chen2021learning} & C & DD & $1\rm{M}$ & 64.79$\pm$5.53 & 87.47$\pm$4.68 \\
        LBC~\cite{chen2020learning} & C & DD & $157\rm{K}$ & 30.97$\pm$4.17 & 55.01$\pm$5.14 \\
        CILRS \cite{codevilla2019exploring} & C & DD & $720\rm{K}$ & 7.47$\pm$2.51 & 13.40$\pm$1.09  \\ 
        % \hline
        VLM + GPT-4 & C & KD & $11\rm{K}$ & 81.31$\pm$2.37 & 94.22$\pm$3.18 \\ 
        \rowcolor[HTML]{ECECEC}\textbf{{\methodname} (w/o Town05)} & C & KD & $11\rm{K}$ & 75.73$\pm$1.36 & 92.10$\pm$1.44 \\
        \rowcolor[HTML]{ECECEC}\textbf{{\methodname}} & C & KD & $11\rm{K}$ & \textbf{83.11$\pm$0.28} & \textbf{94.98$\pm$0.54} \\
        \Xhline{1pt}
    \end{tabular}
    \label{tab:carla-leaderboard}
    \vspace{-5pt}
\end{table*}

\subsection{Evaluation in Closed-Loop Driving}
\label{exp: close-loop}
We conduct closed-loop experiments in CARLA, a popular and realistic open-source simulator, to evaluate the performance of our {\methodname}. 
To validate the effectiveness, we conduct a comprehensive assessment in a closed-loop driving scenario on the Town05 benchmark. Our evaluation metrics include Driving Score (DS), Route Completion (RC), and Infraction Score (IS). RC signifies the proportion of the route successfully navigated by the agent, while IS indicates penalties incurred from accidents. By multiplying RC by IS, we obtain the final metric DS for evaluating our method's driving performance on a route.
Table~\ref{tab:carla-leaderboard} compares our method with competitive methods on the Town05 Short benchmark. 
Specifically, we provide three different configurations to evaluate our methods comprehensively: \emph{\romannumeral 1 )}  VLM + GPT-4 represents directly using non-fine-tuned GPT-4 as the decision module along with the VLM (Qwen-VL \cite{bai2023qwen}); \emph{\romannumeral 2 )} {\methodname} (w/o Town05) represents the dual-process system with the memory bank of 9K samples accumulated from various towns (01-04, 06, 07, and 10) except Town05; \emph{\romannumeral 3 )} {\methodname} denotes the dual-process system with the memory bank of 18K samples accumulated from various towns (01-07, and 10) and 0.1K reflection data in Town05.

As shown in Table~\ref{tab:carla-leaderboard}, our {\methodname} outperforms all other methods that rely solely on camera sensor input. Besides, our method surpasses TransFuser~\cite{chitta2022transfuser}, which additionally utilizes LiDAR sensor inputs.
It is worth noting that in all experiments, we only used a total of $11\rm{K}$ data to fine-tune the VLM,  while all other methods employ tens to hundreds of times more data.
Moreover, our dual-process decision module does not involve human annotations, demonstrating the labeling efficiency of {\methodname}.
For instance, although InterFuser~\cite{shao2023safety} achieves higher performance than our {\methodname}, it relies on 3 million camera and LiDAR annotations, approximately 272 times more than our method.
As shown in the results, we observe that even without prior driving experience in Town05, {\methodname} can surpass other camera-input methods. However, there remains a gap compared to VLM + GPT-4, which integrates the scene understanding module and the {\slowsystem}, achieving a DS of 81.31. This demonstrates that GPT-4's understanding of world knowledge and common sense aids in performing driving-specific tasks. But using GPT-4 directly as the decision-making module is both time-consuming and expensive, making it impractical for deployment in vehicles. 
On the other hand, with the continuous accumulation and adaptation of experience in the test town, our {\methodname} surpasses VLM + GPT-4, while only using a 1.8B model in {\fastsystem}. By leveraging an enriched memory bank, it achieves a final DS of 83.11. This result fully validates the effectiveness of our dual-process decision-making module.
{Moreover, we provide the evaluation results on the Town05 Long benchmark and visualizations in Appendix~\ref{sec:other_exp} and \ref{sec:visualization}.}

\subsection{Ablation Study}
\label{exp: Ablation}
We conduct extensive ablation studies about the number of few shots, size of the memory bank, reflection mechanism, and accumulated experience in a closed-loop driving setup to demonstrate the generalization and continuous learning capabilities of our {\methodname}.

\begin{wrapfigure}{r}{0.39\textwidth}
    \vspace{-20pt}
    \centering
    \includegraphics[width=\linewidth]{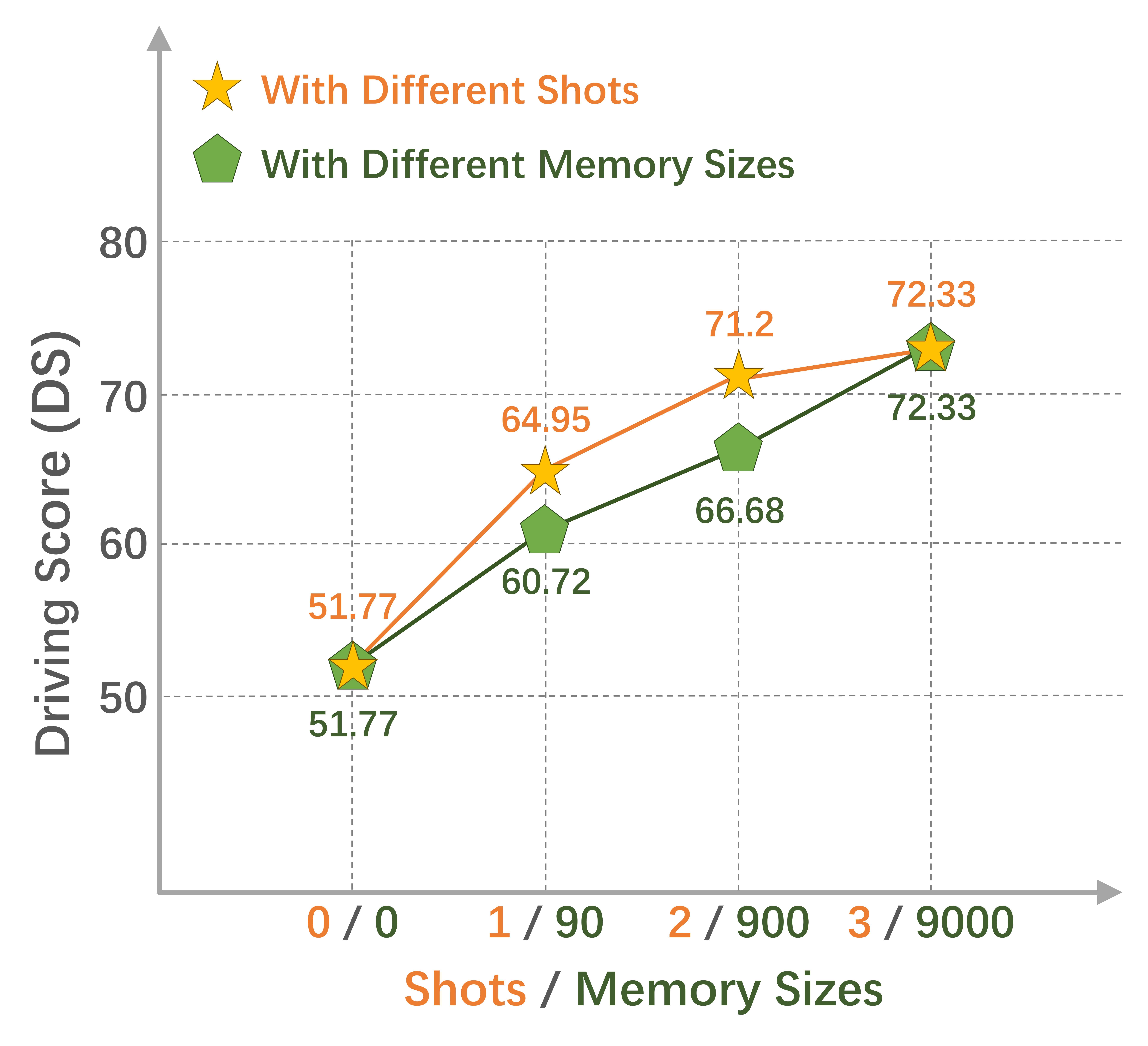}
    \caption{The illustration for ablation studies of few-shot and memory size. See Appendix~\ref{sec:other_exp} for the detailed data.}
    \label{fig:few-shot memsize}
    \vspace{-20pt}
\end{wrapfigure}

\paragraph{Ablation on the number of few-shot.}
We highlight the importance of few-shot prompting in using accumulated experience to guide current decision-making.
The experiments are tested on the Town05 Short benchmark with the memory bank consisting of $9\rm{K}$ samples automatically collected by {\slowsystem} in Town05. The results are presented in Figure~\ref{fig:few-shot memsize}, showing that our approach surpasses several methods (e.g., CLIRS~\cite{raffel2020exploring}, LBC~\cite{chen2020learning} in Table \ref{tab:carla-leaderboard}) even with a zero-shot setting.
Furthermore, there is a notable performance improvement when moving from zero-shot to one-shot scenarios. Moreover, there is a consistent increase in closed-loop experiment results as the number of shots is increased to three, which experimentally demonstrates the value of the experience in the memory bank and the effectiveness of the few-shot strategy.

\begin{table*}[t]
\setlength{\tabcolsep}{5.2pt}
\centering
\renewcommand{\arraystretch}{1.1}
\small
\caption{Generalization of accumulated knowledge in the memory bank. We evaluate the performance of our {\methodname} across various towns (Town05, Town01, and Town04) by leveraging memory banks accumulated from diverse sources. $L_{avg}$ denotes the average lengths of routes within each town.}
\begin{tabular}{c|c|cccc|ccc} 
    \Xhline{1pt}
    \multirow{2}{*}{Test town}
    & \multirow{2}{*}{$L_{avg}$(m)}
    & \multicolumn{2}{c}{Memory (Town01-04, 06)} 
    & \multicolumn{2}{c|}{Memory (Town05)} 
    & \multirow{2}{*}{DS $\uparrow$} & \multirow{2}{*}{RC $\uparrow$}
        & \multirow{2}{*}{IS $\uparrow$}  \\
    & & Few-shot & SFT & Few-shot & SFT & & & \\
    \Xhline{1pt}
    \multirow{5}{*}{Town05} & \multirow{5}{*}{70.1}
    &  & \checkmark  & & &  66.40 & 90.40 & 73.81  \\ 
    &  & \checkmark & \checkmark & & & 75.73 & 92.10 & 82.66 \\
    & &  &  &  & \checkmark & 69.90 & 91.79 & 76.64 \\
    &  &  & & \checkmark & \checkmark & 78.07 & 91.69 & 85.89 \\ 
    & & \checkmark & \checkmark & \checkmark & \checkmark & 83.11 & 94.98 & 87.78  \\ \hline
    Town01
    & 129.1 & & & \checkmark & \checkmark & 68.68 & 100.0 & 68.68  \\ \hline
    Town04
    & 119.3 & & & \checkmark & \checkmark & 95.08 & 97.96 & 96.56\\
    \Xhline{1pt}
\end{tabular}
\label{table:Generalization}
\vspace{-10pt}
\end{table*}

\paragraph{The impact of memory sizes.}
\label{sec: size of memory}
The memory bank contains accumulated experiences crucial for improving the performance of our approach. Therefore, we conduct additional ablation studies to explore the impact of the size of the memory bank, which is equipped with the few-shot strategy (defaulting to 3-shot). We conduct closed-loop evaluations with memory banks of different sizes: \emph{\romannumeral 1 )} base memory bank with $9\rm{K}$ samples. \emph{\romannumeral 2 )} compressed memory banks with 900 and 90 samples evenly sampled from the base memory bank. \emph{\romannumeral 3 )} no memory bank in system.
The quantitative results are presented in Figure ~\ref{fig:few-shot memsize} and illustrate a gradual performance increase as the memory size grows. This further demonstrates the continuous learning capability of our proposed {\methodname}, indicating that our model's performance can improve with accumulated experience. 

\begin{wrapfigure}{r}{0.5\textwidth}
    \vspace{-5pt}
    \centering
    \includegraphics[width=\linewidth]{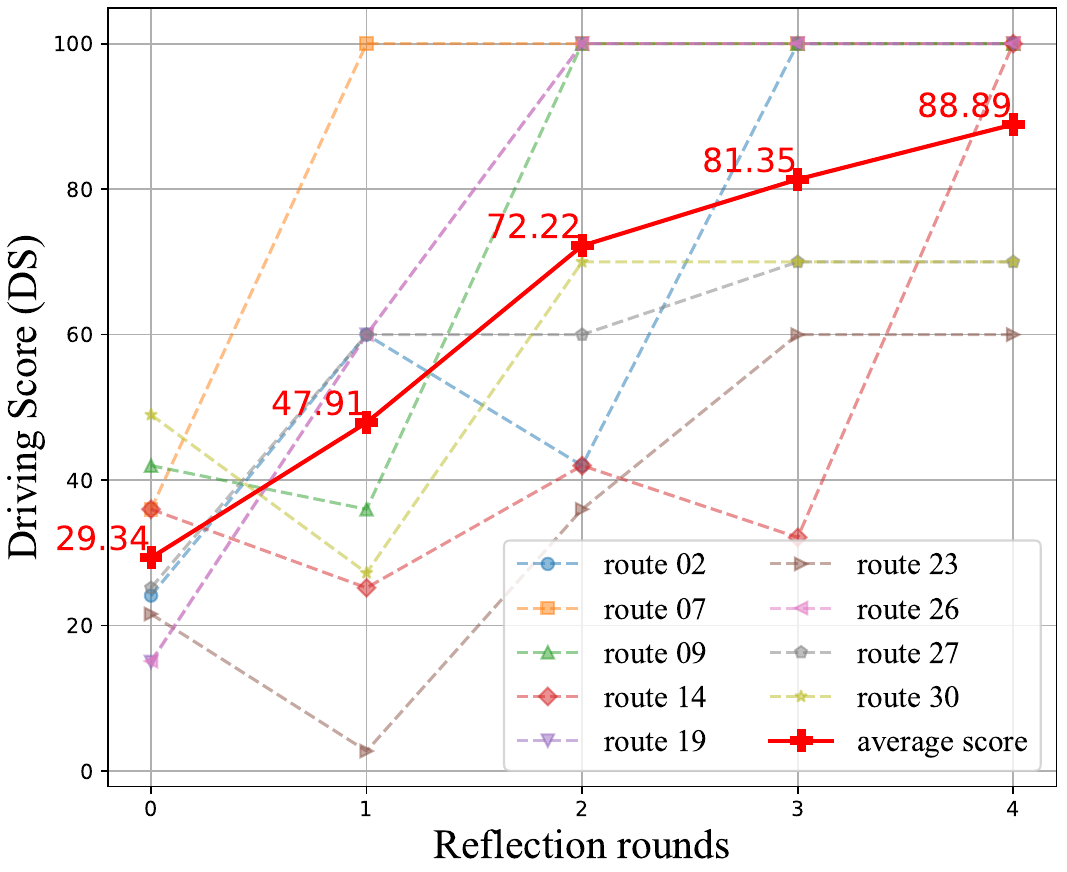}
    \caption{Effectiveness of the reflection mechanism. The $x$-axis represents the rounds of reflection, while the $y$-axis denotes the resulting driving score. The dashed line illustrates performances on different routes after multi-round reflection, and the red "average score" denotes the mean performance across all routes.}
    \label{fig:reflection_rounds}
    \vspace{-5pt}
\end{wrapfigure}

\paragraph{Effectiveness of the reflection mechanism.}
Reflection plays a crucial role in the continuous learning capability of our proposed {\methodname}. It includes reflecting on mistakes and incorporating correct experiences into the memory bank, fostering self-improvement through proactive summarization and accumulation of experiences in unfamiliar scenes, notably corner cases. To enhance experimental efficiency, 
we employ a configuration of three shots and a memory bank with 900 stored memories in the Town05 environment. Reflective experiments are conducted by selecting routes with scores below 50.
incorporated, and the post-reflection experiences were added to the memory bank. 
Figure~\ref{fig:reflection_rounds} indicates that reflection significantly enhances out method's performance, albeit with some instances where individual sequence scores temporarily decrease, potentially due to the inherent randomness of the simulation environment and the limitations of the VLM model.
Addressing these limitations will be the focus of our future work.

\paragraph{Generalization of accumulated knowledge.}
\label{ab:generalization}

Finally, we conduct a series of experiments to demonstrate the generalization and transferability of the experience in the memory bank. The results are illustrated in Table~\ref{table:Generalization}, which shows the robustness of our proposed {\methodname} on different towns. 
As shown in Row 1 and Row 3, solely employing experience adaption (SFT) from different towns with zero-shot, our {\methodname} has achieved commendable performance on the Town05 Short benchmark, surpassing many methods in Table \ref{tab:carla-leaderboard}. 
Implementing the few-shot strategy yields substantial enhancements (Row 2 \& 4), further highlighting the effectiveness of the experience in the memory bank. 
Furthermore, we perform cross-validation to demonstrate the generalization of the accumulated experience, focusing on two setups: \emph{\romannumeral 1 )} testing on Town05 with the memory bank collected from other towns (Row 1 \& 2); \emph{\romannumeral 2 )} testing on the other towns (Town01, 04) with the memory bank accumulated only from Town05 (Row 6 \& 7). 
Comparing Row 2 with Row 4, our {\methodname} utilizing different memory sources achieves comparable performance in the same town (Town05). Additionally, our proposed {\methodname} demonstrates good performance across different towns when using the same memory bank (Row 4, 6 \& 7).
In this section, several low-quality samples in the memory bank on routes with low driving scores are withdrawn when accumulating experience, and no special memory processing is performed on those experiments in Figure~\ref{fig:few-shot memsize}.
\vspace{-3pt}

\section{Conclusion}
In this paper, we introduce {\methodname}, a dual-process closed-loop autonomous driving system with continuous learning, adapting, and improving capabilities. Similar to human attention, our approach selectively prioritizes critical objects that can influence driving decisions, simplifying the scene description and reducing decision-making complexity. Furthermore, the dual-process decision-making module mimics human cognitive processes through a fast, empirical {\fastsystem} and a slow, rational {\slowsystem}. Through reflection mechanisms and a transferable memory bank, {\methodname} continuously improves from past experiences in a closed-loop environment, demonstrating continuous learning capabilities and strong adaptability to various driving scenarios. Moreover, {\methodname} can be seamlessly integrated with the mainstream cloud-edge architectures employed in intelligent vehicles. The {\fastsystem} operates at the edge, enabling instant decision-making within the vehicle, while the {\slowsystem} handles more complex scenarios in the cloud.
\vspace{-3pt}

\section{Limitations and Broader Impacts} \label{limitations}
%limitations

Currently, {\methodname} relies solely on single-frame camera inputs, without any temporal input. Another bottleneck of our approach is the VLM's inability to participate in the reflection mechanism, which hinders further system improvements. Additionally, there is a notable gap between predefined agent behaviors in the CARLA benchmark and those in real-world scenarios, underscoring the need for a high-fidelity world simulator. For impacts, autonomous driving systems gather extensive data on driving behaviors, routes, and passenger movements, raising concerns about data privacy and legal implications. Nonetheless, advancements in technology and regulatory frameworks can help address these issues, paving the way for safer, more efficient, and accessible driving systems.

\section{Acknowledgments}
This research was supported by the NSFC 62088101 Autonomous Intelligent Unmanned Systems. Additionally, it received support from the Shanghai Artificial Intelligence Laboratory, the National Key R\&D Program of China (Grant No. 2022ZD0160104), and the Science and Technology Commission of Shanghai Municipality (Grant No. 22DZ1100102).

{\small
\bibliographystyle{ieeetr}  % abbrv, named
\bibliography{ref}
}

\newpage
%%%%%%%%%%%%%%%%%%%%%%%%%%%%%%%%%%%%%%%%%%%%%%%%%%%%%%%%%%%%
\section*{Appendix}
\appendix

\section{Low-level Control}
\label{sec:low-level control}
As mentioned, our dual-process decision-making module outputs meta-actions (e.g., "AC", "DC", "IDLE", "STOP"), which are further refined to control signals such as steering, acceleration and brake. For the sake of simplicity, we define "AC" to mean acceleration of 1m/s, and "DC" to mean deceleration of -1m/s.

\subsection{Planned Waypoints}
The default route waypoints provided by CARLA are sparse, with distances between consecutive waypoints reaching up to several dozen meters. This makes it difficult for our lower-level controller to decompose meta-actions into control signals. 
To address this issue, we leverage the high-definition map to densify these sparse waypoints into 1-meter interval path-points, which form the reference path for the ego vehicle.
Subsequently, our controller employs the Pure Pursuit algorithm~\cite{coulter1992implementation} to track the reference path, ensuring that the ego vehicle remains on the correct road. The selection of the target path-point for tracking is adaptive, depending on the vehicle's speed, with the controller choosing one of the third to seventh path points ahead.

It is worth noting that high-definition maps are not a necessary requirement for our approach. Alternative methods, such as those proposed in DriveCot~\cite{wang2024drivecot} or TransFuser~\cite{chitta2022transfuser}, which utilize separate neural networks to predict the future reference path based on camera images with sparse navigation information, are also compatible with controller design, without affecting our core methodology.

\subsection{PID Controller}
Given the target path-point and the target vehicle speed, ego vehicle's control method employs two independent PID controllers, building upon previous works \cite{chitta2021neat,chitta2022transfuser,wu2022trajectory,jia2023think,wang2024drivecot}. Specifically, the control system contains two separate PID controllers: one longitudinal for throttle and brake and one lateral for steering. 
The longitudinal PID controller, tuned with gains $K_P = 5.0$, $K_I = 0.5$, and $K_D = 1.0$, takes the current speed and desired target speed as inputs and utilizes a 40-frame buffer to compute the throttle and brake values. Meanwhile, the lateral PID controller, tuned with gains $K_P = 1.0$, $K_I = 0.5$, and $K_D = 0.2$, receives the angle difference between the ego vehicle's heading and the vector pointing to the chosen future waypoint as input, and employs a 20-frame buffer to calculate the steering value.

\section{Dataset Annotations}
\label{sec:data_anno}

\begin{figure}[tbp]
    \centering
    \includegraphics[width=0.9\linewidth]{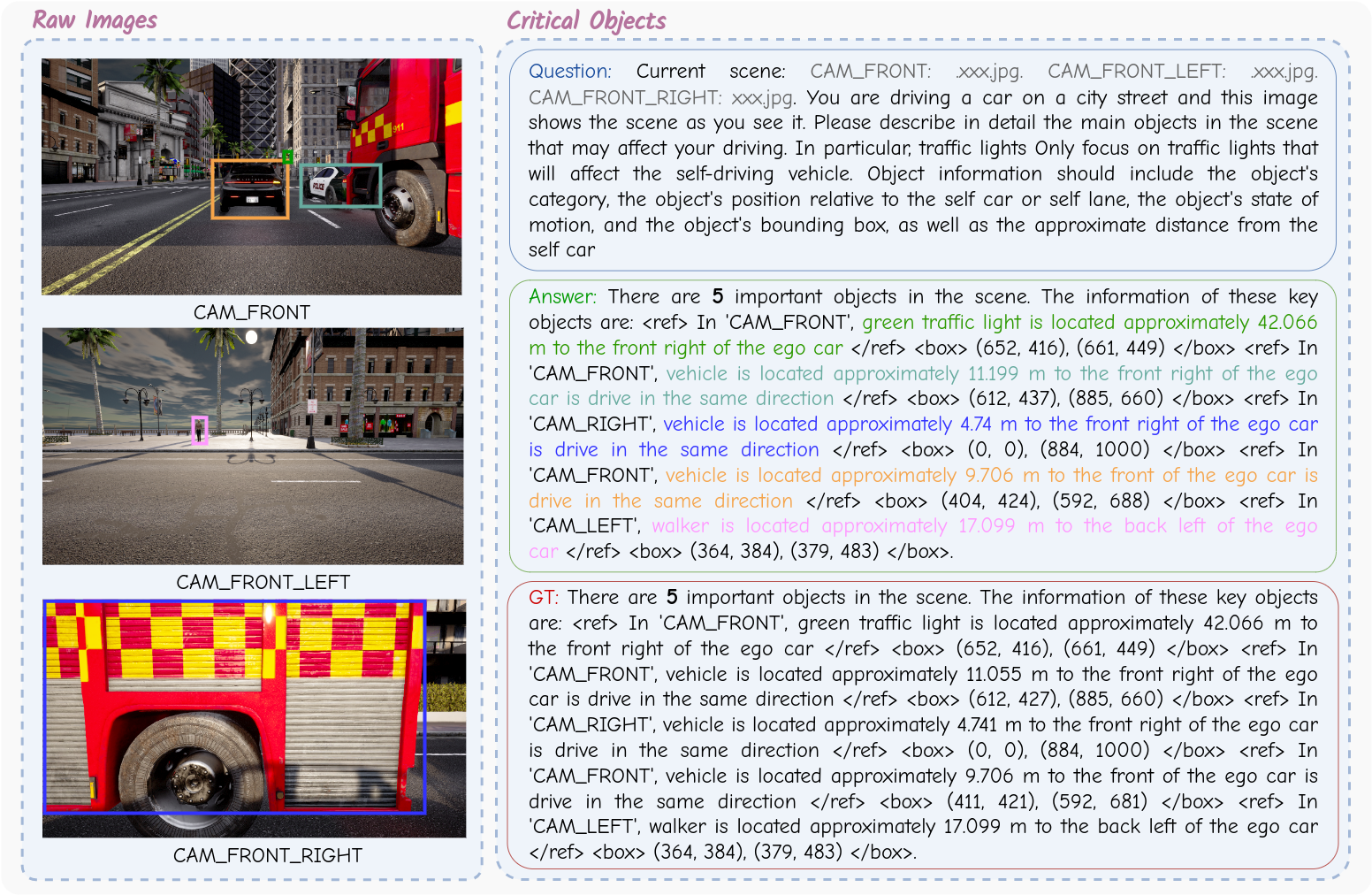}
    \caption{Annotation format and example of scene description generated by VLM on a self-collected CARLA simulation dataset}
    \label{fig:carla}
    \vspace{-10pt}
\end{figure}

\begin{figure}[tbp]
    \centering
    \includegraphics[width=0.9\linewidth]{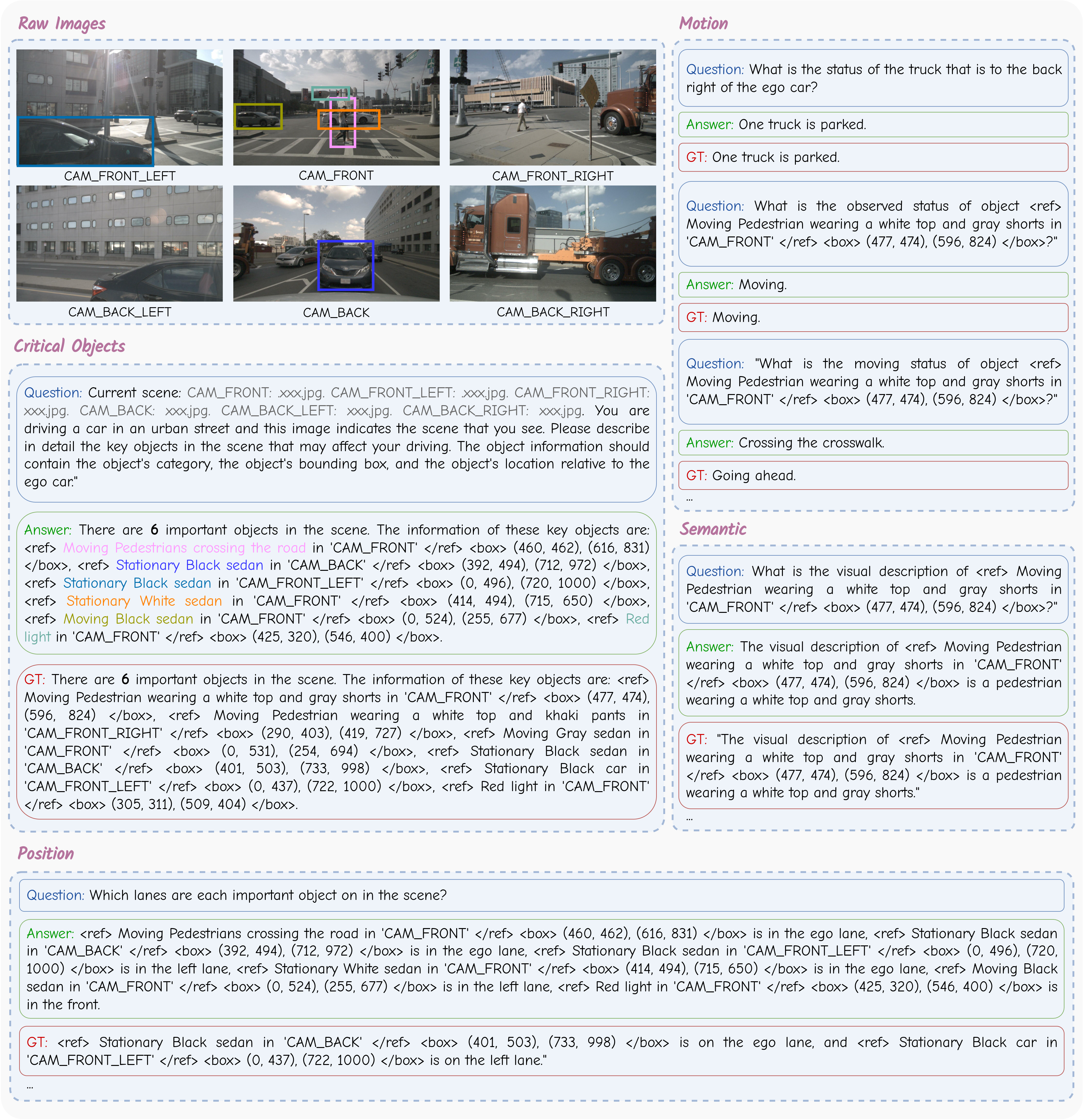}
    \caption{Annotation format and example of scene description generated by VLM on DriveLM dataset.}
    \label{fig:drivelm}
    \vspace{-10pt}
\end{figure}

\begin{figure}[tbp]
    \centering
    \includegraphics[width=\linewidth]{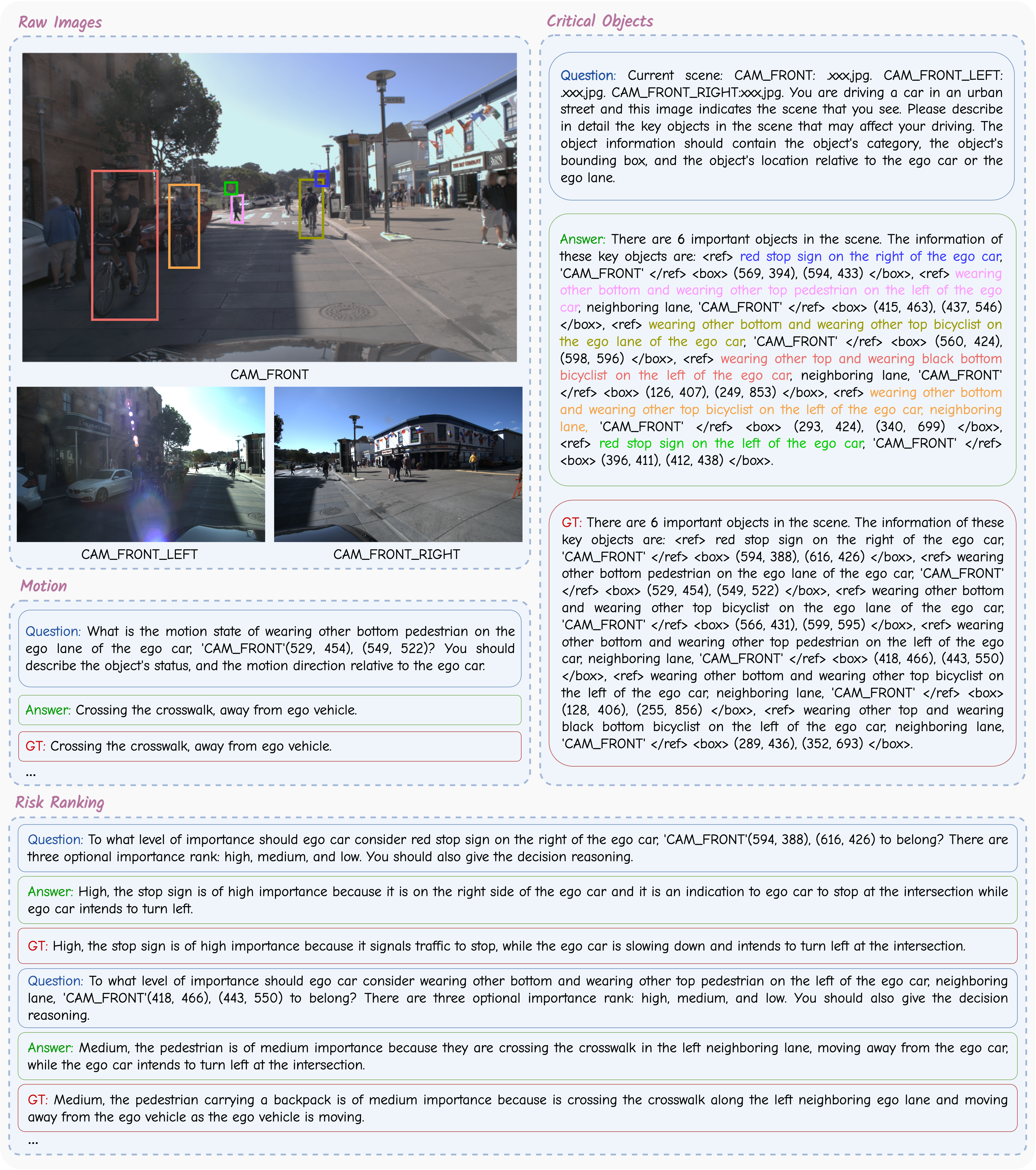}
    \caption{Annotation format and example of scene description generated by VLM on Rank2Tell dataset. It demonstrates the robust performance of our VLM in real-world scenarios}
    \label{fig:rank2tell}
\end{figure}

\begin{figure}[tbp]
    \centering
    \includegraphics[width=0.9\linewidth]{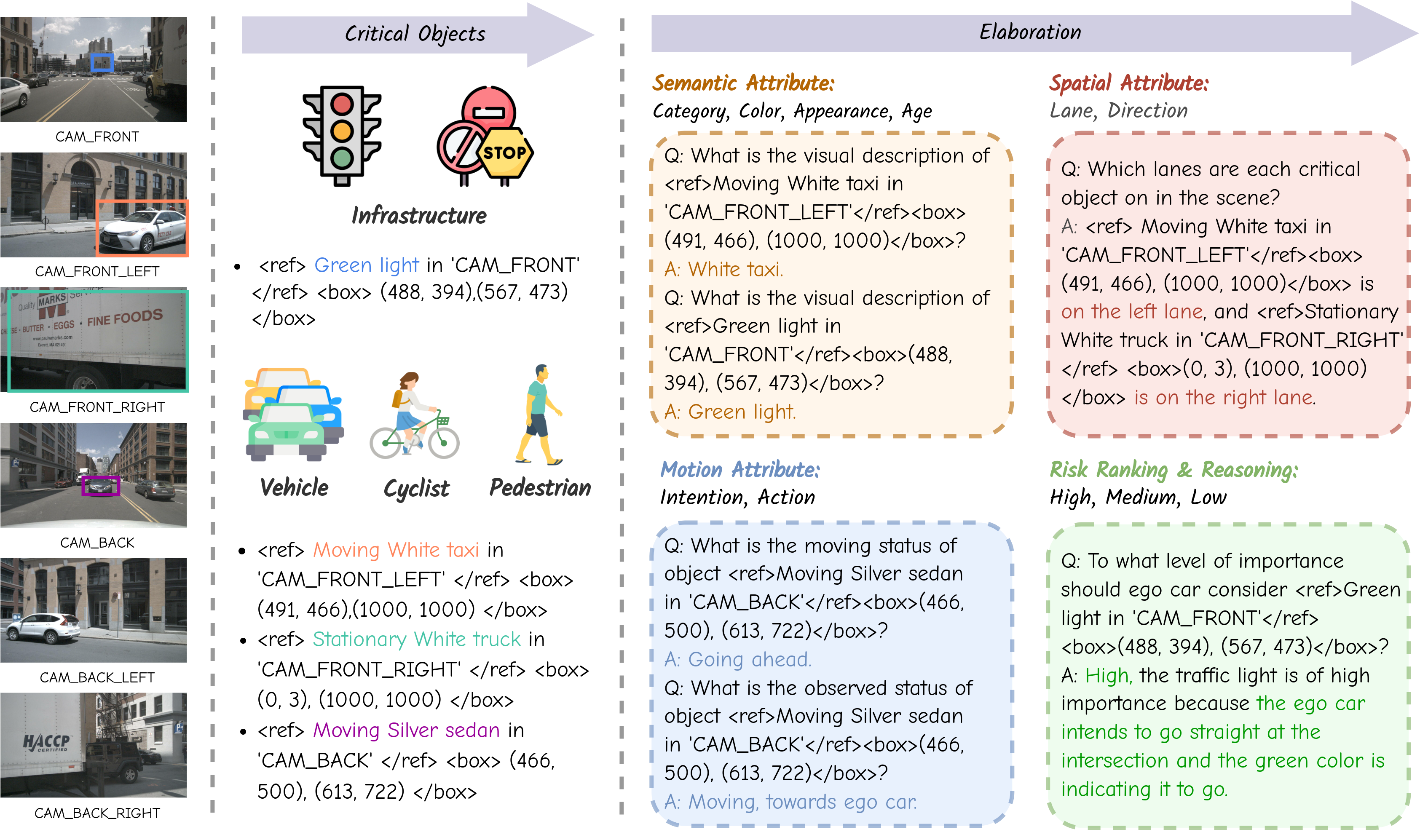}
    \caption{Detailed data format for the descriptions of critical objects.}
    \label{fig:data_illustrate}
\end{figure}

In this section, we present examples of datasets utilized to train the VLM, including the self-collected CARLA dataset (Figure~\ref{fig:carla}), the DriveLM dataset (Figure~\ref{fig:drivelm}), and the Rank2Tell dataset (Figure~\ref{fig:rank2tell}). 
The data annotations principally categorize the critical objects based on four attributes (semantic, spatial, motion, and risk ranking), as outlined in Figure~\ref{fig:data_illustrate} and Section~\ref{method:vlm}. 
It is worth noting that not every dataset comprehensively covers questions and answers pertaining to these attributes. For instance, the DriveLM dataset lacks the selection of important objects; the Rank2Tell dataset merges semantic and spatial information into a single overview. 

To conduct closed-loop experiments, we specifically define questions about the semantic and spatial details of key objects in the CARLA dataset and design several automatic annotation rules.
Specifically, the critical objects are defined as those situated in the front, left, and right fields of view: \emph{\romannumeral 1 )} Vehicles and cyclists positioned within 20 meters of the ego car or less than 60 meters in the ego lane. \emph{\romannumeral 2 )} Pedestrians within 40 meters of the ego car. \emph{\romannumeral 3 )} Traffic lights that control the vehicle's travel direction and stop signs. 

In addition to presenting the datasets, we further elaborate on the description generated by the VLM. The examples described in Figure~\ref{fig:carla}, \ref{fig:drivelm}, \ref{fig:rank2tell} demonstrate VLM’s proficiency in accurately identifying key objects that influence driving decisions and effectively describing the related attributes.
We observe that VLM not only excels in simulated environments but also demonstrates robust performance in real-world scenarios. The versatility of our scene understanding module enables the proposed {\methodname} with the promise of practical applicability across real-world scenarios.

\section{Prompt Details}
\label{sec:prompt}
\begin{figure}[htbp]
    \centering
    \includegraphics[width=\linewidth]{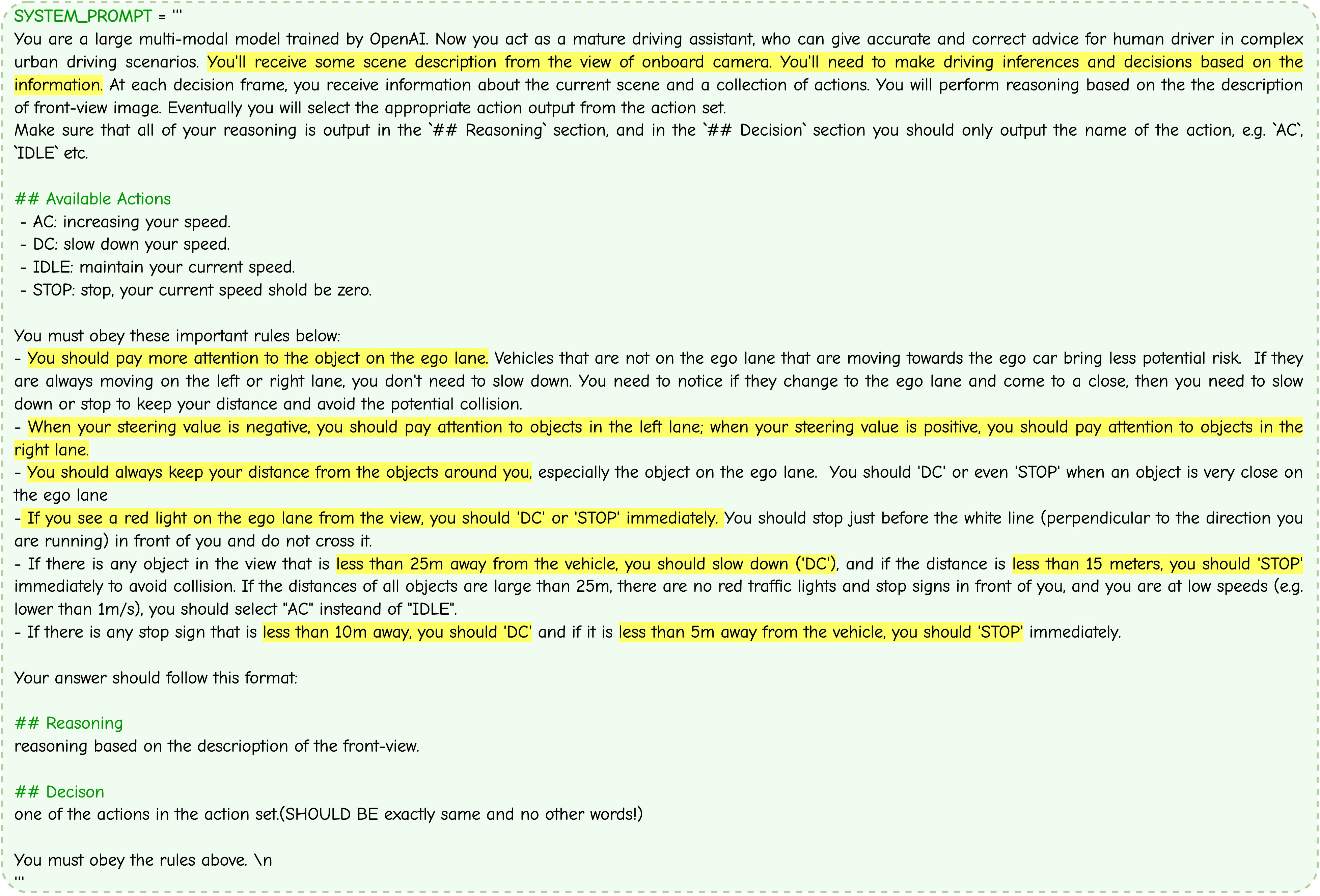}
    \caption{System prompt for \slowsystem}
    \label{fig:system prompt}
\end{figure}

\begin{figure}[htbp]
    \centering
    \includegraphics[width=0.93\linewidth]{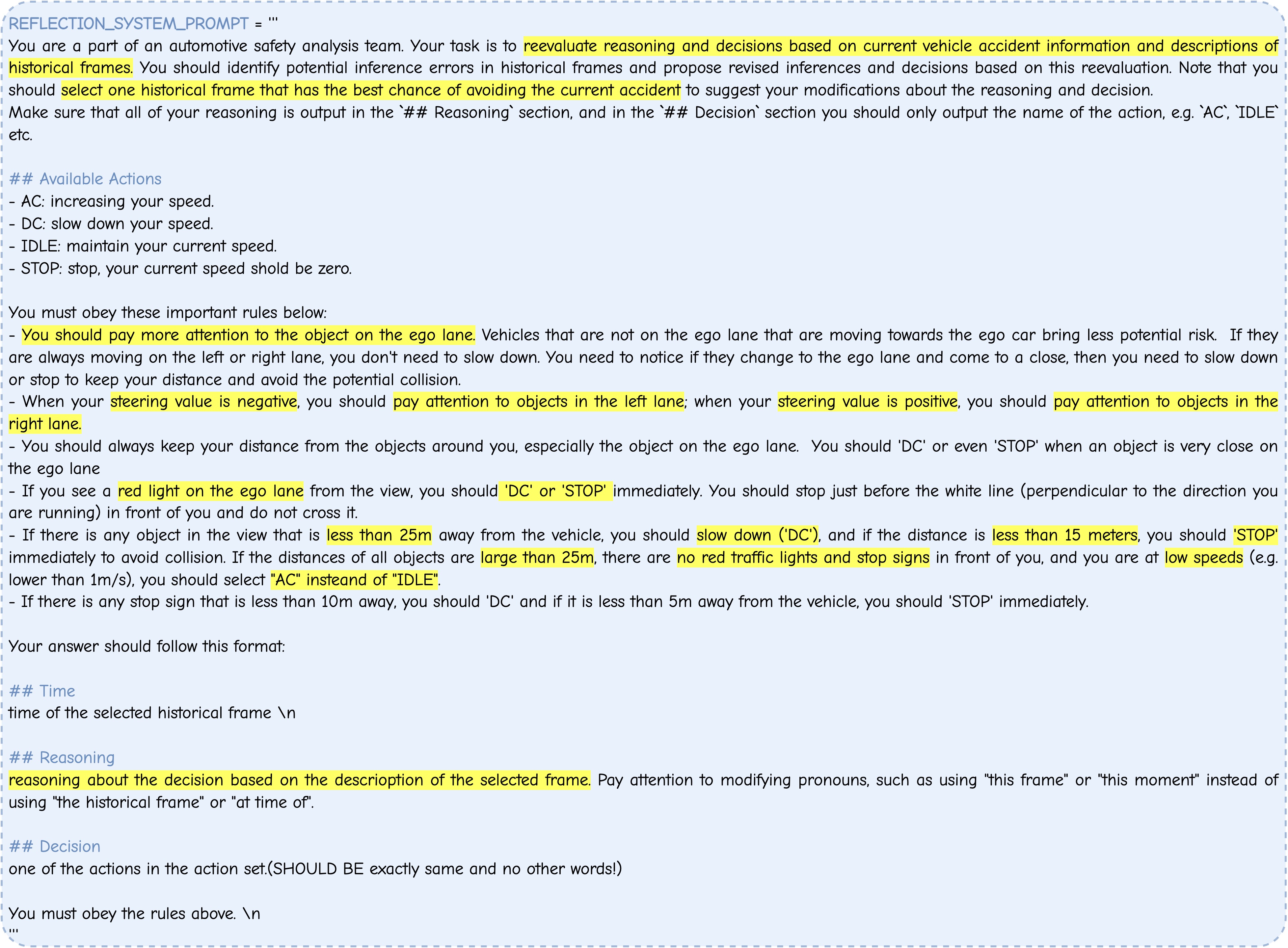}  
    \caption{System prompt in the reflection mechanism}  
    \label{fig:reflection prompt}
\end{figure}

\begin{figure}[htbp]
    \centering
    \includegraphics[width=0.93\linewidth]{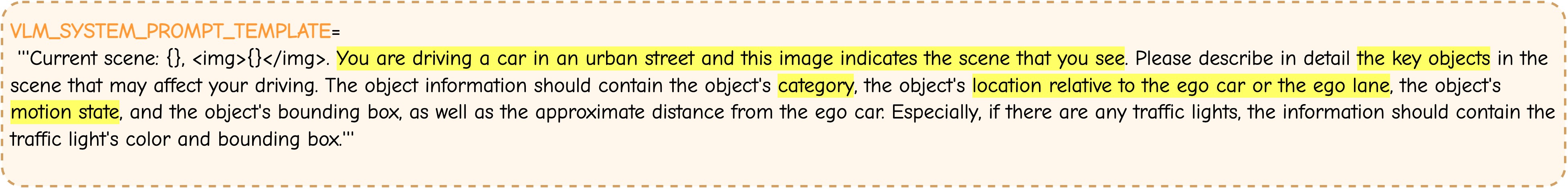}  
        \caption{VLM system prompt} 
        \label{fig:vlm prompt}
\end{figure}

We outline the specifics of the system prompt (Figure~\ref{fig:system prompt}) utilized by our {\slowsystem} during the accumulation of experience within a closed-loop environment. The prompt consists of task definitions, meta-actions, adherence to traffic rules, and the desired output format. Furthermore, the figure illustrates the system prompt (Figure~\ref{fig:reflection prompt}) utilized during the reflection procedure, which contains fundamental descriptions from the previous prompt alongside criteria for identifying potential errors within historical frames. We also detail the prompts (Figure~\ref{fig:vlm prompt}) of the VLM utilized for identifying the critical objects in the traffic scenes.

\vspace{-10pt}

\begin{table*}[htbp]
    \centering
    \small
    \caption{Comparison of our {\methodname} with competitive methods in Town05 Long benchmark. "DD" and "KD" denote data-driven and knowledge-driven, respectively.}
    \begin{tabular}{l|cccccc}
    \hline
        % \multirow{2}{*}{Method} & \multirow{2}{*}{Modality} & Annotations & \multirow{2}{*}{Driving Score (\%) $\uparrow$} & \multirow{2}{*}{Road Completion (\%) $\uparrow$} \\ 
        % & & & & \\\hline
        {Method} & {Modality} & Type & Annotations & {DS (\%) $\uparrow$} & {RC (\%) $\uparrow$} & {IS (\%) $\uparrow$} \\ \hline
        DriveMLM~\cite{wang2023drivemlm} & L+C & DD & $ 2\rm{M}$ & 76.1 & 98.1 & 78.0 \\
        ThinkTwice~\cite{jia2023think} & L+C & DD & $2\rm{M}$ & 70.9 & 95.5 & 75.0 \\ 
        InterFuser~\cite{shao2023safety} & L+C & DD & $3\rm{M}$ & 68.3 & 95.0 & 72.0 \\ 
        TransFuser~\cite{chitta2022transfuser} & L+C & DD & $ 228\rm{K}$ & 31.0 & 47.5 & 77.0 \\
        \hline
        VAD~\cite{jiang2023vad} & C & DD & 228$\rm{K}$ & 30.3 & 75.2 & - \\
        TCP~\cite{wu2022trajectory} & C & DD & $420\rm{K}$ & 57.2 & 80.4 & 73.0 \\
        NEAT~\cite{chitta2021neat} & C & DD & $130\rm{K}$ & 37.7 & 62.1 & 61.0 \\
        Roach~\cite{zhang2021end} & C & DD & - & 43.6 & 80.4 & 54.4  \\
        WOR~\cite{chen2021learning} & C & DD & $1\rm{M}$ & 44.8 & 82.4 & 54.0 \\
        LBC~\cite{chen2020learning} & C & DD & $157\rm{K}$ & 7.1 & 32.1 & 22.1 \\
        CILRS \cite{codevilla2019exploring} & C & DD & $720\rm{K}$ & 3.7 & 7.2 & 51.4  \\ \hline
        {\methodname} (ours) & C & KD & $11\rm{K}$ & 51.7 & 100 & 51.7 \\
        \hline
    \end{tabular}
    % \vspace{-20pt}
    \label{tab:carla-long-leaderboard}
\end{table*}

\begin{table*}[htbp]
\centering
\begin{minipage}{.5\linewidth}
    \centering
    \small
    \caption{Ablation on the number of few shots.}
    \setlength{\tabcolsep}{3.8pt}
    \begin{tabular}{c|ccc} 
    \hline
    Shots & DS (\%) $\uparrow$ & RC  (\%) $\uparrow$
    & IS  (\%) $\uparrow$ \\
    \hline
    0
    & $51.77$ & $99.72$  & $52.45$ \\
    1
    & $64.95$ & $100.0$ & $64.95$ \\
    2
    & $71.20$ & $94.86$ & $73.60$ \\
    3
    & $\textbf{72.33}$ & $\textbf{100.0}$ & $\textbf{72.33}$ \\
    \hline
    \end{tabular}
    \label{table:few-shot}
\end{minipage}%
\begin{minipage}{.5\linewidth}
    \centering
    \small
    \caption{Affect of the memory size for few-shot learning.}
    \setlength{\tabcolsep}{3.8pt}
    \begin{tabular}{c|ccc} 
    \hline
    Mem Sizes & DS (\%) $\uparrow$ & RC  (\%) $\uparrow$
    & IS  (\%) $\uparrow$ \\
    \hline
    0
    & $51.77$ & $99.72$  & $52.45$ \\
    90
    & $60.72$ & $98.30$ & $62.30$ \\
    900
    & $66.68$ & $97.93$ & $68.10$ \\
    9000
    & $\textbf{72.33}$ & $\textbf{100.0}$ & $\textbf{72.33}$ \\
    \hline
    \end{tabular}
    \label{table:memory-bank}
\end{minipage}
% \vspace{-20pt}
\end{table*}

\section{Reflection mechanism}
\label{sec:reflection}
As mentioned in Section \ref{method:slow}, we employ the {\slowsystem} to reflect on traffic accidents, facilitating the system’s capabilities to continuously improve. Specifically, in the reflection mechanism, we maintain a memory queue $\boldsymbol{Q} = \{(D_i, R_i, S_i)\}_{i=0}^9$, storing the previous 10 samples from {\fastsystem} at a frequency of 1 Hz. When an accident occurs, samples in $\boldsymbol{Q}$ are fed to {\slowsystem} for reflection. This prompts the system to identify keyframes that caused the accident and provide correct reasoning and decisions, which are then added to the memory bank. The detailed prompt used in the reflection procedure is depicted in Figure \ref{fig:reflection prompt}. In addition, Figure~\ref{fig:reflectioncase} depicts a specific reflection procedure.
\vspace{-10pt}

\section{Other experiments}
\label{sec:other_exp}
{Moreover, we provide the evaluation results on the Town05 Long benchmark in Table~\ref{tab:carla-long-leaderboard}. The results show that our {\methodname} can still achieve competitive results compared to those methods that take images as input, despite using annotated data that is much smaller. }
{Our study delves into the nuanced effects of few-shot learning and memory size through rigorous ablation analyses, as detailed in Tables~\ref{table:few-shot} and~\ref{table:memory-bank}. The baseline performance of the SFT model without memory is presented in the first row, revealing a clear positive correlation with increased shots and expanded memory capacity in closed-loop experiments. Table~\ref{table:few-shot} substantiates the efficacy of the few-shot heuristic process and highlights the success of our method in identifying similar memories. Table~\ref{table:memory-bank} demonstrates the incremental performance gains as memory accumulates, exemplifying the principle of continuous learning.}

% We also added additional ablation studies to explore the influence of Qwen-VL and LLaVA for scene understanding. This includes scene understanding evaluation results on the Rank2tell (real) and CARLA (simulated) datasets, detailed below. We provided Grounded scores, including precision, recall, and F1 score, to assess the models' grounding performance, as well as Chat Scores, including language score (ROUGE) and GPT score (GPT-4-turbo), to evaluate the models' reasoning and question-answering capabilities. From these results, it is evident that Qwen-VL exhibits superior grounding abilities.
\begin{figure}[htbp]
    \centering
    \includegraphics[width=0.86\linewidth]{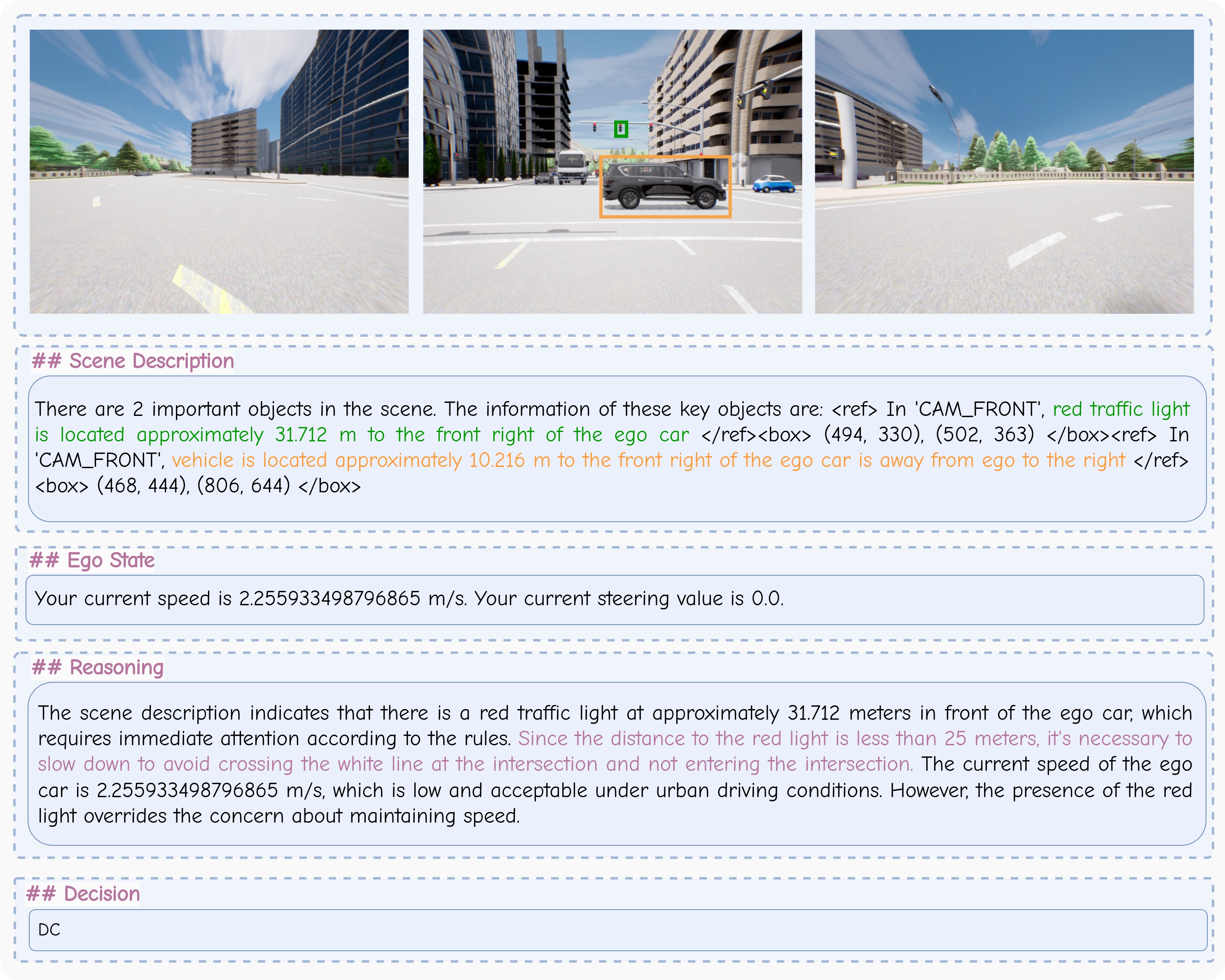}
    \caption{Case study at the intersection.}
    \label{fig:case_1}
\end{figure}

\begin{figure}[htbp]
    \centering
    \includegraphics[width=0.86\linewidth]{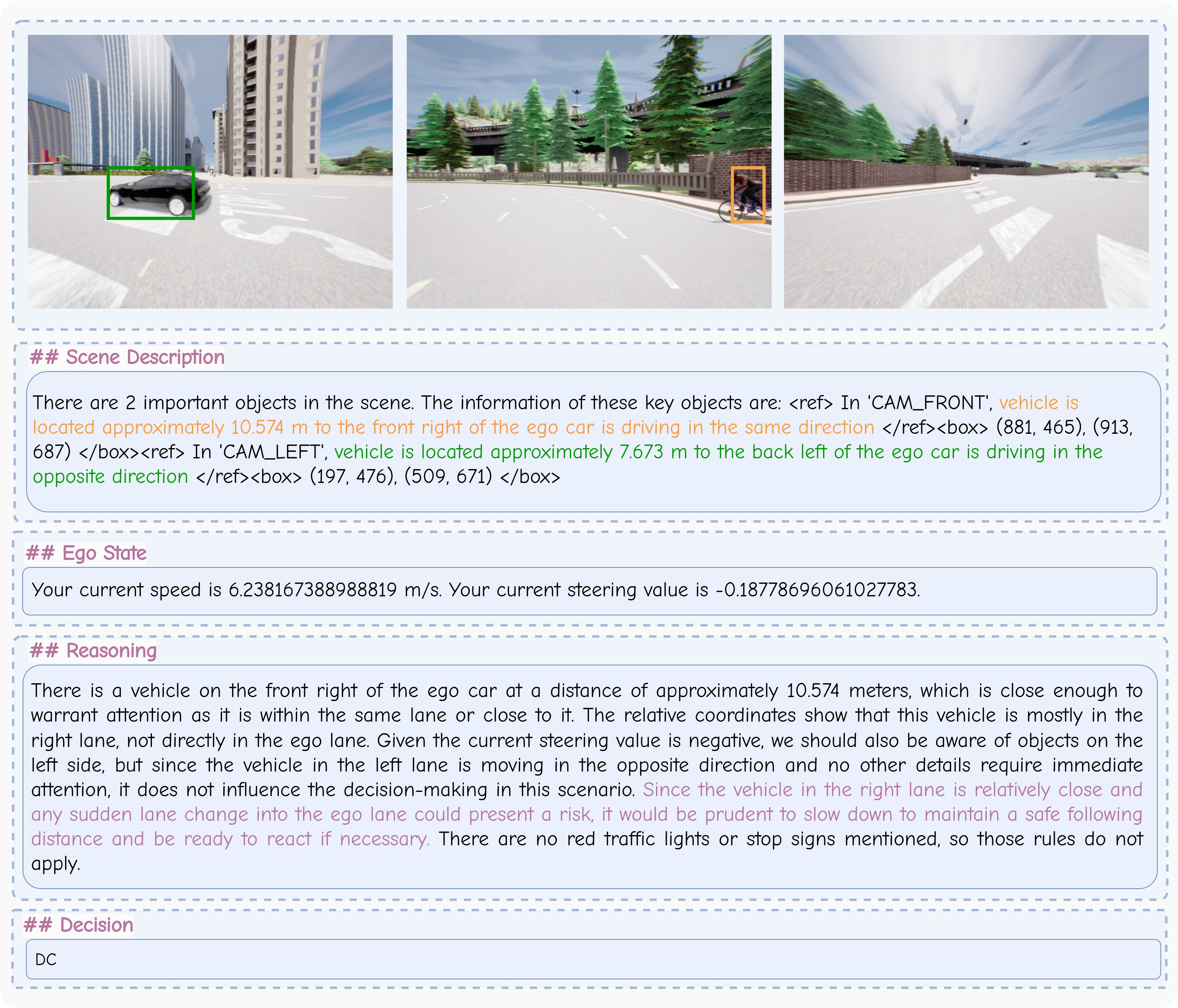}
    \caption{A case where a bicycle suddenly appears.}
    \label{fig:case_2}
\end{figure}

\clearpage
\section{Visualization Cases}
\label{sec:visualization}
\paragraph{Representative cases.} We also provide several examples to show the qualitative performance of our proposed {\methodname}, as illustrated in Figure~\ref{fig:case_1}, \ref{fig:case_2}. These cases present the zero-shot results of our proposed system: The VLMs take the images of the current scene and generate the scene descriptions, which are fed into the decision module (\fastsystem) along with the ego state (i.e., speed and steering values) for reasoning and decision-making. 
As shown in Figure~\ref{fig:case_1}, VLM correctly perceives the important traffic participants at the intersection, such as the red lights and passing vehicles, prompting the decision module to make a deceleration decision. Figure~\ref{fig:case_2} showcases a challenging scenario where a cyclists suddenly appears from the curb while the vehicle is moving. The system adeptly detects the risk and promptly responds with a timely deceleration.

\begin{figure}[htbp]
    \centering
    \includegraphics[width=0.98\linewidth]{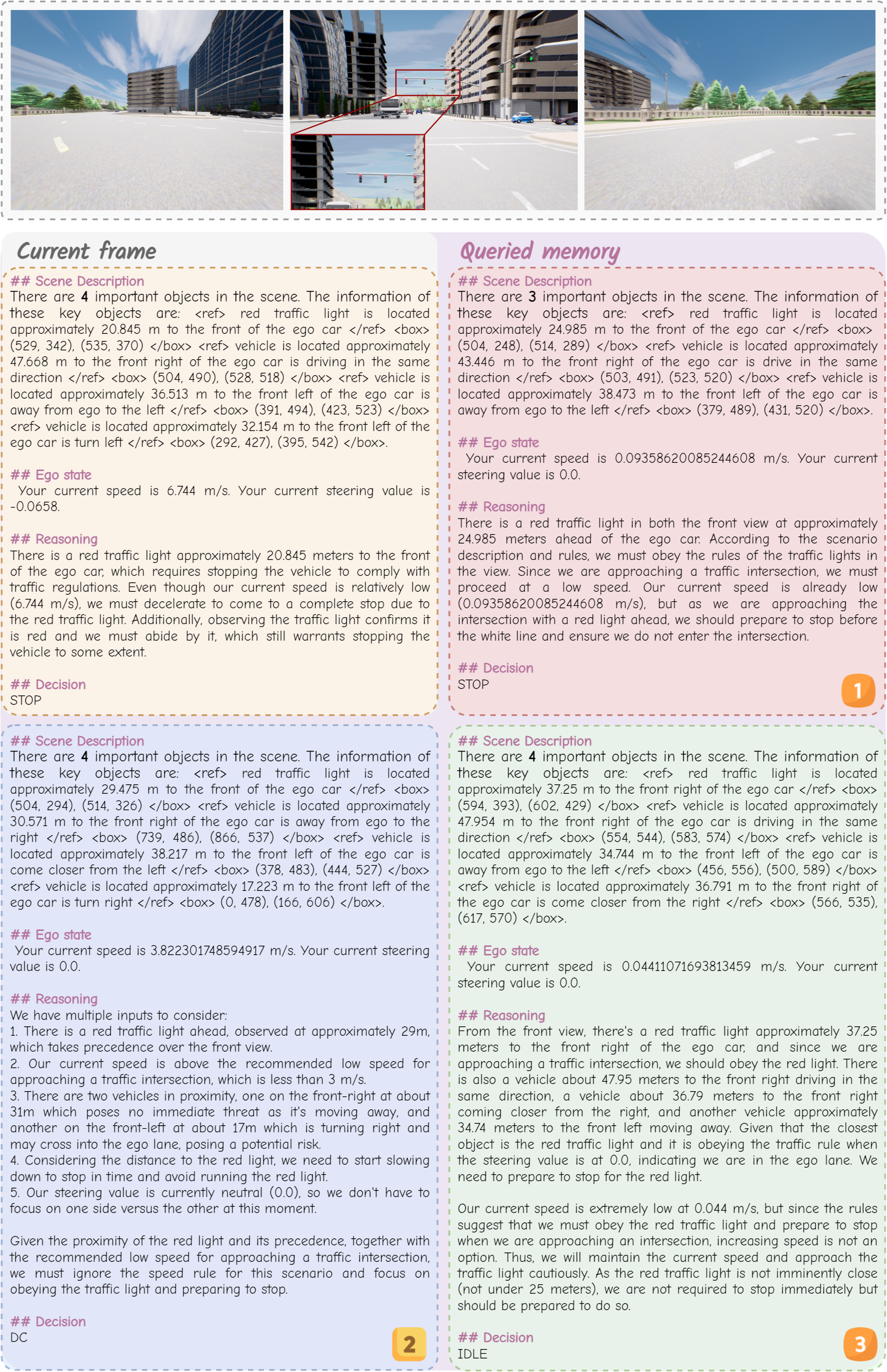}
    \caption{Case study for few-shot strategy.}
    \label{fig:fewshotcase1}
\end{figure}

\paragraph{Cases of few-shot prompting.} Figure~\ref{fig:fewshotcase1} presents a case to show the results using the few-shot strategy. As mentioned in Section~\ref{method:fast}, several memory samples are queried based on similarities of the scene descriptions' embedding between the current scene and those in the memory bank. From this case, we can see that the queried samples are highly related to the current scene. And our system {\methodname} observes there is a red traffic light and several vehicles in the front. According to the previous experience and current ego status, the system suggests that it should obey the red traffic light and prepare to stop. 

\begin{figure}[htbp]
    \centering
    \includegraphics[width=0.98\linewidth]{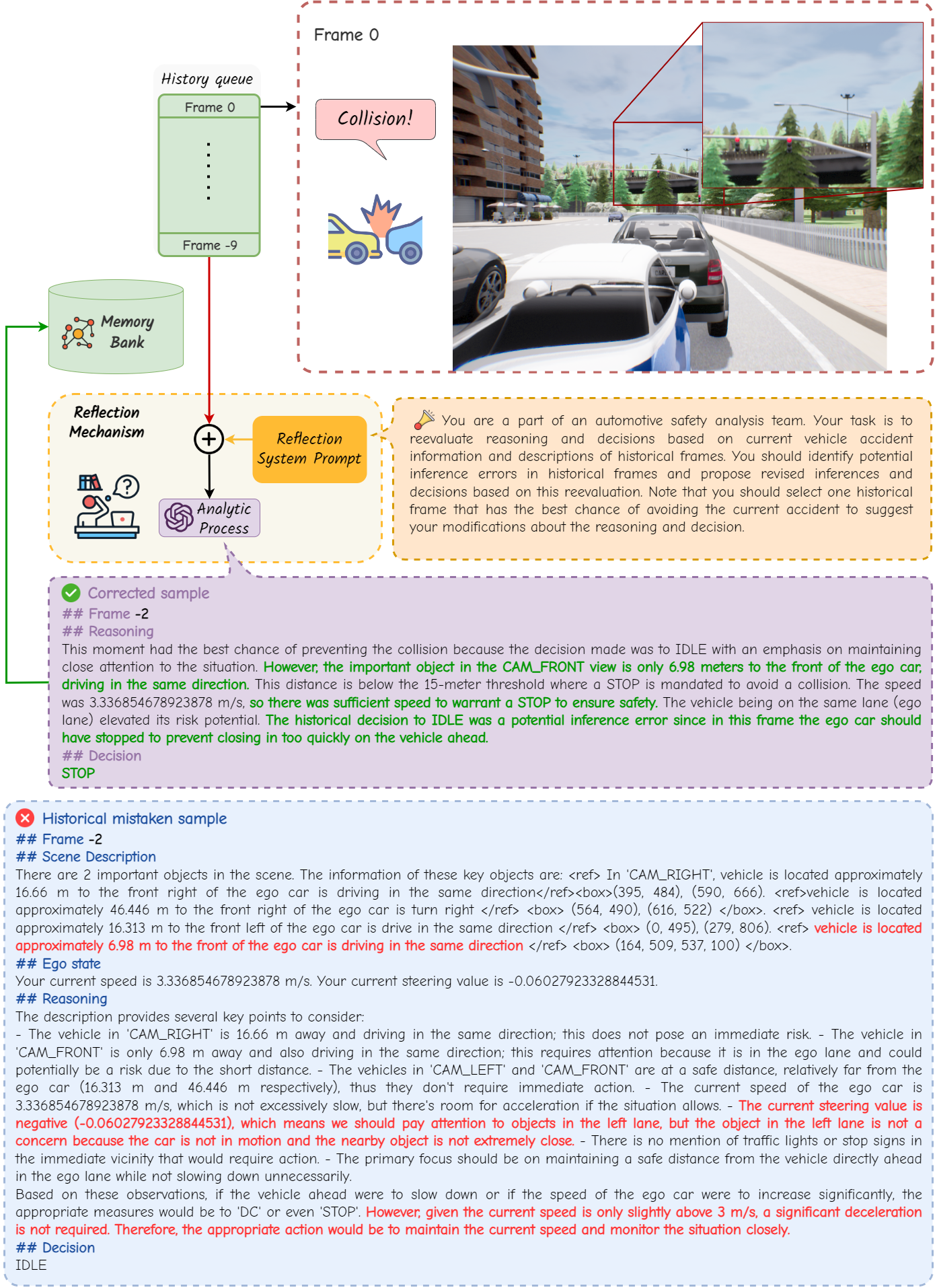}
    \caption{Case study for reflection mechanism.}
    \label{fig:reflectioncase}
\end{figure}

\paragraph{Cases of reflection mechanism.}
We also provide an example to illustrate the reflection mechanism in Figure \ref{fig:reflectioncase}. When an incident occurs, the historical descriptions, reasoning, and decisions are fed into the {\slowsystem} to analyze the sequence of events meticulously, identify potential errors, and provide the reflected reasoning and decision. As the case shows, after the collision occurs at the current frame (frame 0), the {\slowsystem} found there exist reasoning errors at the previous frame (frame -2). For instance, when a vehicle is dangerously close (6.98m) ahead of the ego car, the initial analysis by the {\fastsystem} incorrectly identifies the crucial object and misinterprets its movement status, necessitating immediate attention. Furthermore, the {\fastsystem} struggles to comprehend the relationship between the ego car's speed and the safe distance from the vehicle in front, resulting in poor decision-making. Conversely, during the reflection procedure, the {\slowsystem} accurately identifies the vehicle 6.98 meters ahead as the primary concern. It questions the prior decision to maintain an "IDLE" status, suggesting it may have been an inference error, as stopping would have been necessary to avoid closing in too rapidly on the vehicle ahead in this scenario.

\begin{figure}[htbp]
    \centering
    \includegraphics[width=0.92\linewidth]{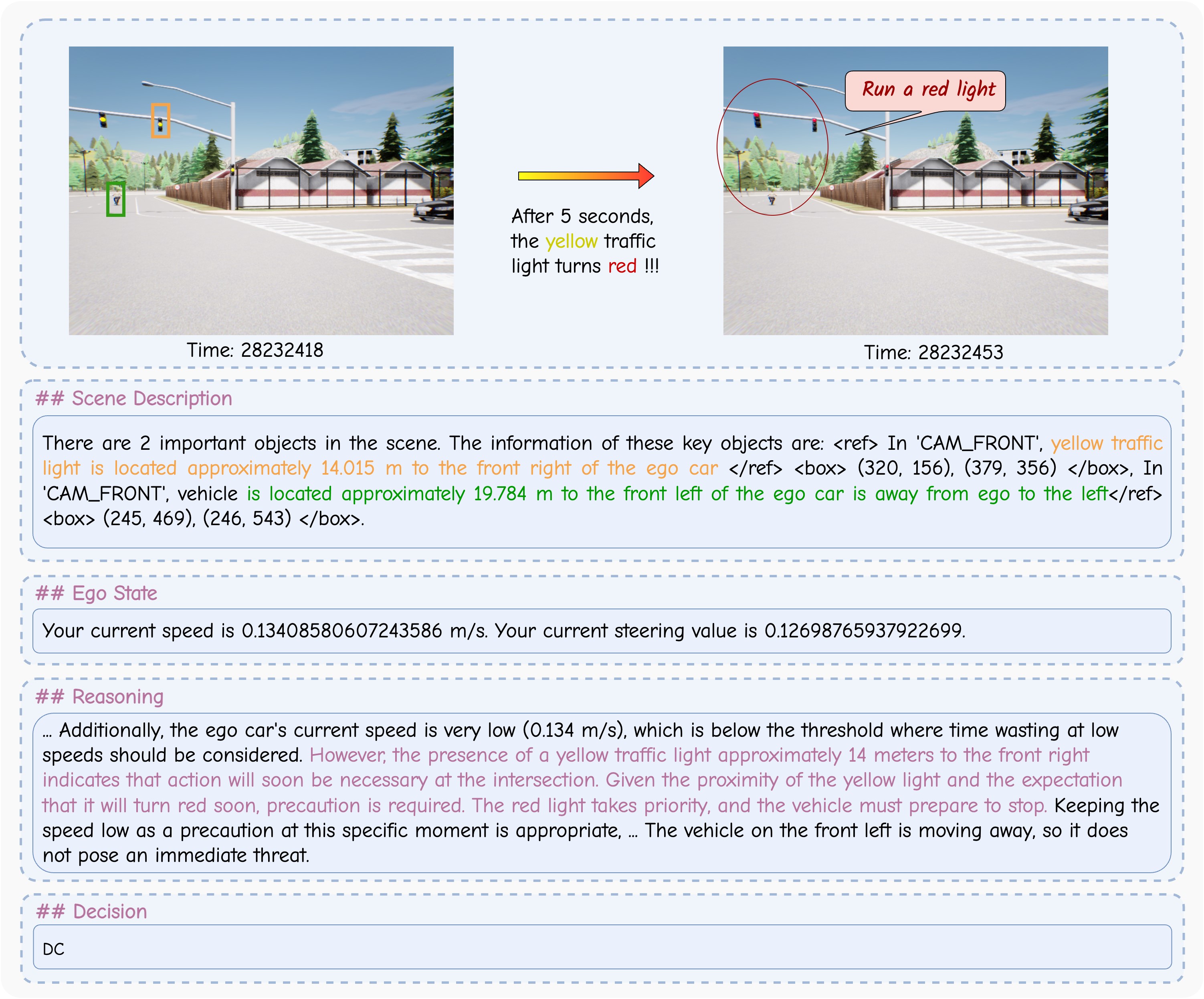}
    \caption{Failure case that running the red light.}
    \label{fig:failurecase1}
\end{figure}

\begin{figure}[htbp]
    \centering
    \includegraphics[width=0.92\linewidth]{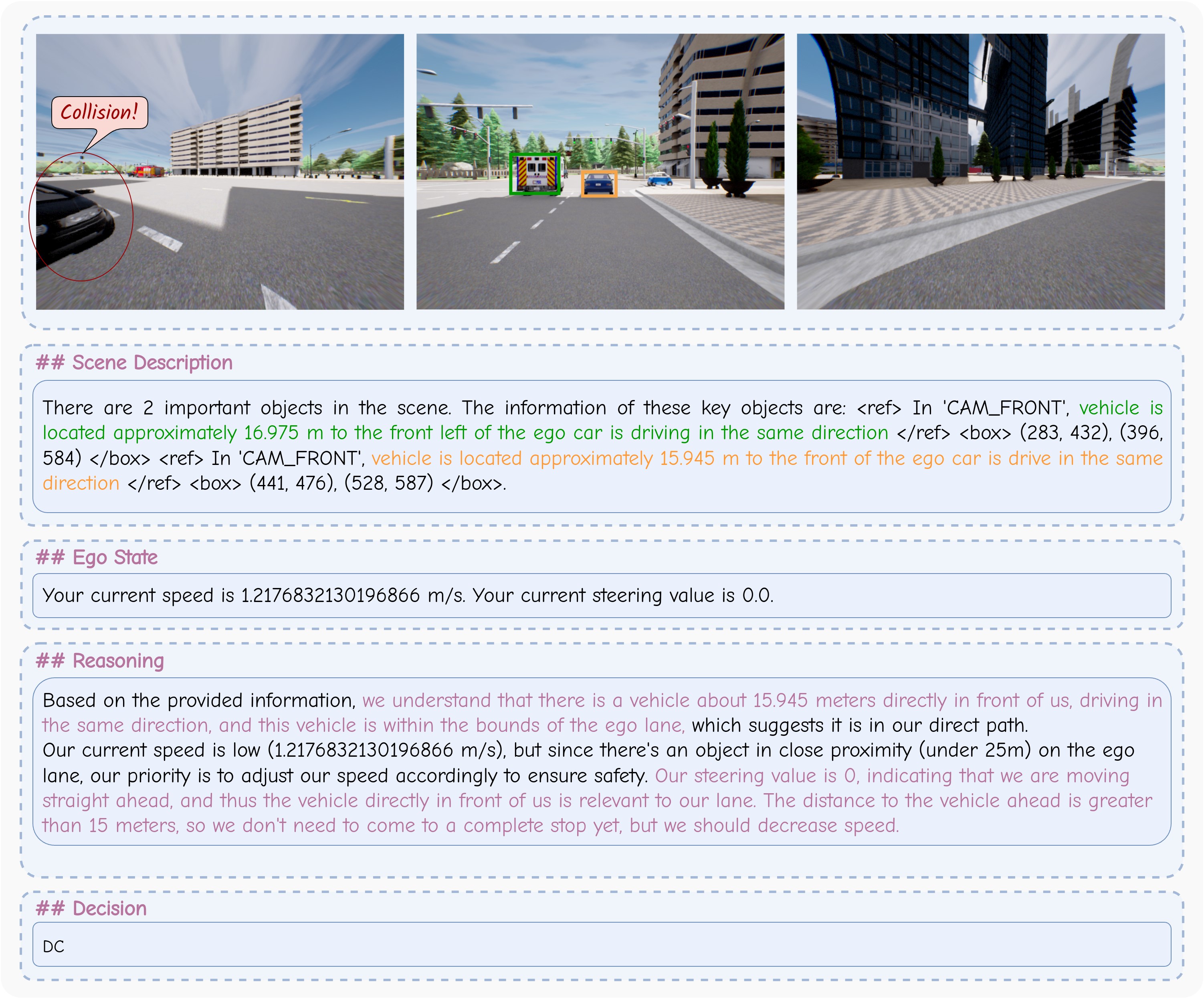}
    \caption{Failure case of collisions caused by other vehicles.}
    \label{fig:failurecase}
\end{figure}

\paragraph{Failure Cases.}
We have included two typical failure cases in the uploaded PDF.  
(1) ``Run a red light” as shown in Figure \ref{fig:failurecase1}. In this scenario, the system lacks temporal information regarding the yellow light's remaining duration, making it difficult to determine whether to accelerate through or stop. When the light is yellow, the system cautiously issues a “DC” command, causing the vehicle to cross the stop line slowly.  When the light turned red, CARLA interpreted this as running a red light, even though a “STOP” command was issued at this time.
(2) ``Collision” as shown in Figure \ref{fig:failurecase}. In this case, the VLM did not detect the car at the left rear edge of the field of view due to the camera’s field of view limitation. Furthermore, in the CARLA setting, other vehicles will not proactively yield to the ego vehicle, leading to collisions caused by other vehicles. 

%%%%%%%%%%%%%%%%%%%%%%%%%%%%%%%%%%%%%%%%%%%%%%%%%%%%%%%%%%%%

\end{document}